%% file: Main.tex
\documentclass[english, a4paper, 12pt, twoside]{article} 

\usepackage{subcaption} 
\usepackage{textcomp}
\usepackage[T1]{fontenc, url}
\usepackage[utf8]{inputenc}
\usepackage{titlesec}
\usepackage{multirow}
\usepackage{adjustbox}
\usepackage{graphicx}
\usepackage{mathtools}
\usepackage{amsmath, amssymb, amsthm} % Mathematical packages
\usepackage{parskip} % Removing indenting in new paragraphs
\usepackage{color}
\usepackage{subcaption} 
\usepackage{appendix}
\usepackage{chngcntr} % needed for correct table numbering
\usepackage{hyperref}
\counterwithin{table}{section} % numbering of tables 
\counterwithin{figure}{section} % numbering of figures
\numberwithin{equation}{section} % numbering of equations
\usepackage{xparse,nameref}
\usepackage{algorithm2e} % For algorithms
\RestyleAlgo{ruled}
\usepackage{amsmath}
\usepackage[bottom]{footmisc} % Footnotes are fixed to bottom of page
\usepackage[english]{babel} % document language
\graphicspath{{02_Images/}{../02_Images/}} % path to 02_Images
\usepackage[margin=2.54cm]{geometry} % sets margins for the document
\usepackage{setspace}
\usepackage{microtype}

\titleformat*{\section}{\LARGE\bfseries} % \section heading
\titleformat*{\subsection}{\Large\bfseries} % \subsection heading
\titleformat*{\subsubsection}{\large\bfseries} % \subsubsection heading
\titleformat{\paragraph}
{\normalfont\normalsize\bfseries}{\theparagraph}{1em}{}
\titlespacing*{\paragraph}
{0pt}{3.25ex plus 1ex minus .2ex}{1.5ex plus .2ex}

\usepackage{fancyhdr}
\usepackage{subfiles}
\usepackage{array}
\usepackage[rightcaption]{sidecap}
\usepackage{wrapfig}
\usepackage{float}
\usepackage[labelfont=bf]{caption} % bold text for captions
\usepackage[para]{threeparttable} % fancy tables, check these before you use them
\usepackage{url}
\usepackage[table,xcdraw]{xcolor}
\usepackage{makecell}
\usepackage{hhline}

\usepackage[numbers]{natbib}
\def\biblio{\clearpage\bibliographystyle{unsrtnat}\bibliography{References.bib}} % defines the \biblio command used for referencing in subfiles - DO NOT CHANGE

\fancypagestyle{frontpage}{
	\fancyhf{}

	\vspace*{4\baselineskip}
	
	\fancyhead[C]{ \includegraphics[width=3in]{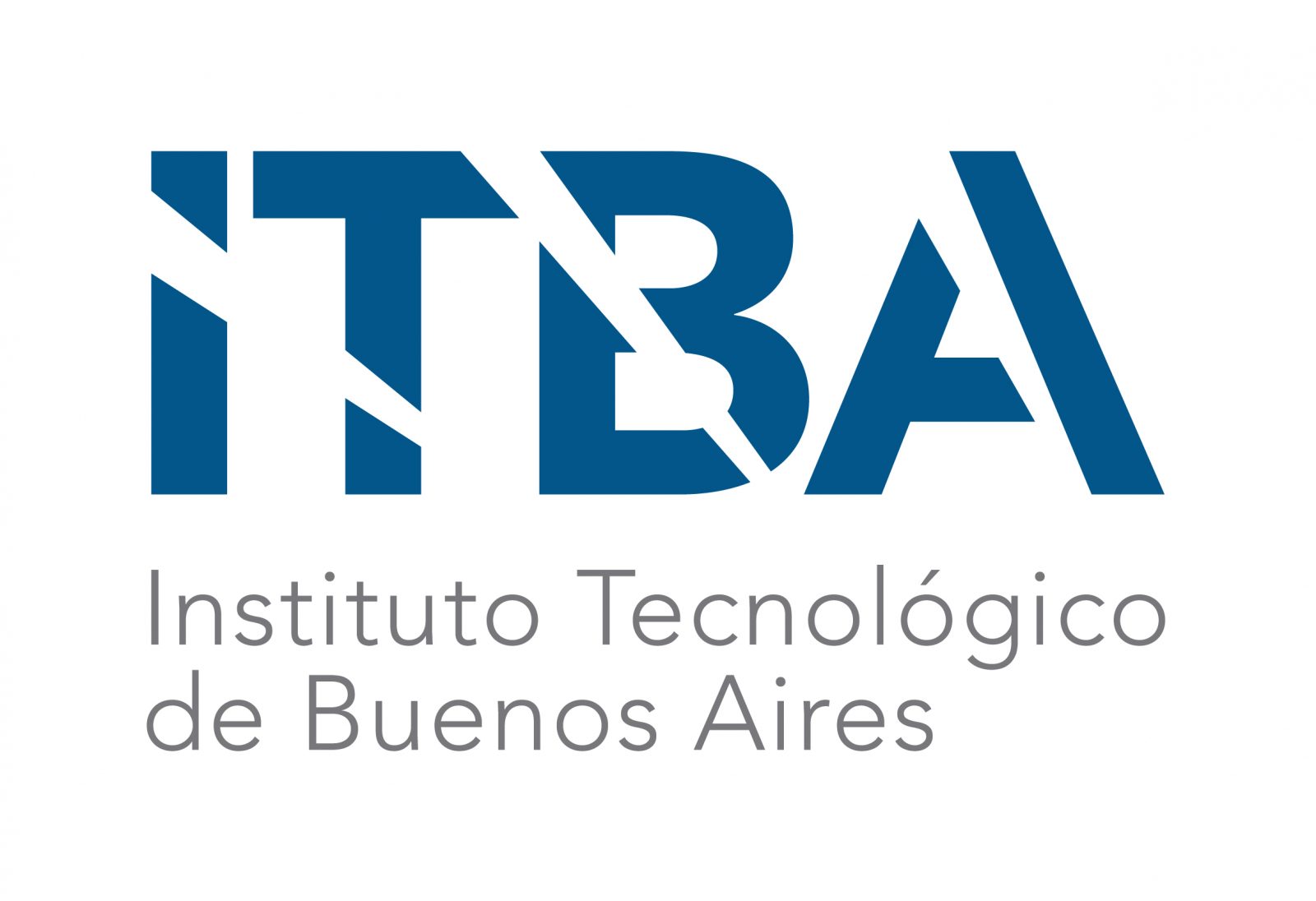}}
}

\begin{document}

\def\biblio{} % resets the biblio command, if not here a new reference list will be produced after every chapter

\include{01_Chapters/000NHH-Frontpage}
\restoregeometry % restores the margins after frontpage
\nocite{*} % uncomment if you want all sources to be printed in the reference list, including the ones which are not cited in the text 

\pagenumbering{gobble} % suppress page numbering
\thispagestyle{plain} % suppress header
\clearpage\mbox{}\clearpage % add blank page

\pagenumbering{roman} % starting roman page numbering
% \newpage
% \section*{Introduction}
%     \subfile{01_Chapters/Introduction}

\newpage
\section*{Abstract}
    \subfile{01_Chapters/00Abstract}

\newpage
\tableofcontents

\newpage
{\setstretch{1.0} 
\listoffigures}
 
\newpage
{\setstretch{1.0} 
\listoftables}

\newpage
\addtocontents{toc}{\protect\setcounter{tocdepth}{4}} % sets depth of toc to 4, 1.1.1.1
\pagenumbering{arabic} % Starting arabic page numbering
\setcounter{page}{1} % sets pagecounter to 1
\setcounter{tocdepth}{0}

\section{Introduction} % section/chapter name

\subfile{01_Chapters/01Introduction} % including the subfile for the chapter
\clearpage % clears the page after the chapter is finished

\section{Review of similar works}
    \subfile{01_Chapters/02ReviewOfSimilarWorks}
\clearpage

\section{Face-frame stabilization in image projection to latent space}
    \subfile{01_Chapters/03FaceFrame}
\clearpage

\section{Materials And Methods}
    \subfile{01_Chapters/04MaterialsAndMethods}
\clearpage
  
\section{Experiments}
    \subfile{01_Chapters/05Experiments}
\clearpage

\section{Results}
    \subfile{01_Chapters/06Results}
\clearpage

\section{Conclusion}
    \subfile{01_Chapters/07Conclusion}
\clearpage

\section{Future Work}
    \subfile{01_Chapters/08FutureWork}
\clearpage

\section{Acknowledgements}
    \subfile{01_Chapters/Acknowledgement}
\clearpage

\newpage
\renewcommand\refname{References} % name for the reference list
{\setstretch{1.0} % linespacing for the references
\addcontentsline{toc}{section}{References} % to change the name of the references in the TOC
\bibliography{References.bib} % adds the references to the document
}

% \newpage
% \renewcommand{\appendixpagename}{Appendix} % Heading of appendix
% \renewcommand{\appendixtocname}{Appendix} % name of appendix in TOC
% \appendixpage 
% \addappheadtotoc

% \begin{appendices}
%     \subfile{01_Chapters/Appendix}
% \end{appendices}

\end{document}

%% file: 01_Chapters/000NHH-Frontpage.tex
\begin{titlepage}
	
	\newgeometry{top=1 in, bottom=1 in, left=1 in, right= 1 in} 
	
	\thispagestyle{frontpage}
	
	\begin{center}
		
		\vspace*{6\baselineskip}

		{\Huge \textbf{Controlling Face's Frame generation in StyleGAN's latent space operations\\}}
		
		\large{\textit{Modifying faces to deceive our memory}}\\
		
        \vspace*{1,5\baselineskip}

		\large{\textbf{Agustín Roca\\Nicolás Britos}}\\
		\large{\textbf{Supervisor: Rodrigo Ramele}}\\
		
		\vspace{1,5\baselineskip}
		
		\large{Final Project}\\
		\large{\textbf{Computer Engineering Degree}}\\
		
		\vspace{1,5\baselineskip}

	\end{center}
	
	\vspace*{6\baselineskip}
	
\end{titlepage}

%% file: 01_Chapters/00Abstract.tex
Innocence Project is a non-profitable organization that works in reducing wrongful convictions. In collaboration with \emph{Laboratorio de Sueño y Memoria} from Instituto Tecnológico de Buenos Aires (ITBA), they are studying human memory in the context of face identification. They have a strong hypothesis stating that human memory heavily relies in face's frame to recognize faces. If this is proved, it could mean that face recognition in police lineups couldn't be trusted, as they may lead to wrongful convictions. This study uses experiments in order to try to prove this using faces with different properties, such as eyes size, but maintaining its frame as much as possible.\\
In this project, we continue the work from a previous project~\citep{lozanoherran} that provided the basic tool to generate realistic faces using StyleGAN2. We take a deep dive into the internals of this tool to make full use of StyleGAN2 functionalities, while also adding more features needed by the Lab, such as projecting a face from a target face image or modifying certain of its attributes, including mouth-opening or eye-opening.\\
As the usage of this tool heavily relies on maintaining the face-frame, we develop a way to identify the face-frame of each image and a function to compare it to the output of the neural network after applying some operations, such as the modification of the eye-opening. The objective is to have a numeric value that measures how much does a certain operation change the face-frame and to know how much of it is maintained, having a clearer perspective of which face image may be a better candidate for generating false memories.\\
We conclude that the face-frame is maintained when modifying eye-opening or mouth opening. When modifying vertical face orientation, gender, age and smile, have a considerable impact on its frame variation. And finally, the horizontal face orientation shows a major impact on the face-frame. This way, the Lab may apply some operations being confident that the face-frame won't significantly change, making them viable to be used to deceive subjects' memories.

\par\vspace*{\fill} % Moves keywords to the bottom of the page
\textbf{\textit{Keywords --}} StyleGAN, GAN, Generative Image Modeling, Innocence Project, False Memories, Laboratorio de Sueño y Memoria, ITBA, Face-Frame, Face Landmarks, Image Segmentation, Image Processing % Add you all the keywords associated with your thesis here

\biblio % Needed for referencing to working when compiling individual subfiles - Do not remove

%% file: 01_Chapters/01Introduction.tex
Understanding how our brains store and recover memories can be really powerful. Marketing, psychology, design, biology, medicine or even computer science can have great breakthroughs by knowing how our brains remember things. Better advertisements~\citep{hafez}, having a more solid base of knowledge to fight Alzheimer~\citep{alzheimer}, dealing with traumas in an efficient way~\citep{hogberg} or creating new models of information based on it~\citep{fukushima}; all these can be potential consequences of understanding memories in our brains. However, we also know that our memory is malleable~\citep{schacter}. It can be manipulated and affected by external factors and deceive ourselves~\citep{loftus}.\\
Having this information, it is natural to doubt the accuracy of human memory. This is the main reason why Innocence Project is doing its research. They are investigating cases where the eyewitnesses were the main evidence for a conviction. They theorize that many of these cases may be result of corruption, manipulation, or honest mistakes when identifying the culprit of the crime. In 1983, around 52\% of wrongful convictions were results of eyewitnesses mistakes~\citep{rattner}.\\
Innocence Project Argentina, partnered with \emph{Laboratorio de Sueño y Memoria} from Instituto Tecnológico de Buenos Aires (ITBA), to research how humans remember faces, which are the key factors for our brains to identify faces and how accurate is human memory for face identification. The experiments the lab is conducting consists on simulating a crime that a subject witnesses. Later, the subject is given the task to try to recognize the perpetrator of the crime in a lineup where the criminal may be present or not. The objective of the experiment is to know which are the attributes that are common between the real criminal and the identified person.\\
During this research, they theorized that the contour of the face, the hair and the ears, are the things that our memory first notice in a face. This set of properties is called the "face-frame".\\
Currently, the Lab is using a tool to generate faces similar to the criminal in the experiment~\citep{lozanoherran}. This tool is based on NVidia's StyleGAN2~\citep{stylegan2}, a Generative Adversarial Network (GAN). A GAN is a deep-learning-based model that allows to create synthetic data which is indistinguishable from real data. Nvidia used this model and changed its architecture to create StyleGAN2, capable of generating realistic images of faces, which is why is used by the Lab.\\
Albeit the tool being used by the Lab is capable of generating images of faces, it's still missing some key functionalities to fully take advantage of StyleGAN2, like being able to change one face's attributes using the very same neural network, or mapping an existing face to a one generated by the network.\\
This way, the project can be divided in two main parts. First, we implement the missing functionalities that StyleGAN2 offers: style mixing and projecting an image into the latent space. These functionalities provide additional tools for the Lab to generate faces, especially when they have specific requirements about how the face they want to generate should look like. In addition, we use some latent directions known by the community~\citep{luxemburg} to be able to modify some attributes of the face. This attributes include eye-opening, mouth-opening, smile, face orientation, presumable age and gender, among others. \\
The second part consists in measuring how much of the face differs from the original one after modifying it with these functionalities. These also allows us to find a way to project an image maintaining its frame as much as possible. Style mixing is not part of the experiments as it only changes color scheme and microstructure of the image~\citep{stylegan1}.\\
In this work we will first present the main stakeholders of this investigation, a brief introduction to human's memory, Generative Adversarial Networks and how faces generated from them can be modified within the same network (specifically in StyleGAN2) and this project's initial status in Section \ref{section:review_similar_works}. In Section \ref{section:face_frame_materials} and \ref{section:materials} readers will find all the definitions and functions we've used to make our measurements, the faces we've used and the setup we have required. Furthermore, in Section \ref{section:experiments} we explain the experiments we have done and in Section \ref{section:results} we present the results we've obtained making a focus on face-frame deviation when modifying faces when moving across some of the neural network directions and in Section \ref{section:conclusion} we discuss the importance and significance of these results, and how they may be used. Finally, in Section \ref{section:future_work} and Section \ref{section:acknowledgement} we conclude this work with future works and acknowledgements.

\biblio % Needed for referencing to working when compiling individual subfiles - Do not remove

%% file: 01_Chapters/02ReviewOfSimilarWorks.tex
\label{section:review_similar_works}

\subsection{Innocence Project}
Innocence Project is one of the 69 organizations that are part of the Innocence Network, that work to free innocent people from jail and to prevent wrongful convictions. This work can be with legal services or investigation of the case presented. ~\citep{innocenceproject}\\
In the United States, the Innocence Project reported over 375 DNA exonerations~\citep{westmeterko}. The causes of wrongful conviction that the organization are aware of are: eyewitness identification, false confessions, forensic science errors, poor defense, among others.~\citep{innocenceproject}\\
Innocence Project Argentina is part of the Innocence Network, and it works in Argentinian cases. This organization is the one that got in touch with \emph{Laboratorio de Sueño y Memoria} from Instituto Tecnológico de Buenos Aires (ITBA).\\
The Lab does research in neuroscience. Some of the areas they are working on are: activation of memories while sleeping, sleep paralysis, wave processing for the brain and false memories formation~\citep{labsym}.

\subsection{Memories}
In 2019, Zlotnik and Vansintjan defined memory as "the capacity to store and retrieve information"~\citep{memdef}. The process of formation of memories have different phases.\\
The first of these phases is the \textbf{acquisition}. In this phase, the sensory stimuli are encoded into neurochemical representations. The second phase is the \textbf{consolidation}, that is a period in which the memory is stabilized in order to subsist through time. The \textbf{retrieval} phase is where the information in a memory can be recovered. A consolidated memory can go through a \textbf{re-consolidation} phase, in which its information can be modified and it is re-stabilized.~\citep{forcatoreconsolidation}\\
This last phase is the one that has a key role for the experiments of the Lab. There are some theories about how false memories are created. Between them, there is the activation-monitoring theory, which explains the creation of false memories using two principles. The first came as a conclusion of an experiment that consists in giving the subject a list of words and later ask if a certain word was in the list or not. It was discovered that if the words in the list were related somehow, and the word asked later followed the same topic, the subject would think the word was in the list, even if it that was not the case. The second principle states that if the person cannot remember the source of the information of interest, that person may create a false memory.~\citep{roediger}\\

\subsection{Generative Adversarial Networks}
A Generative Adversarial Network (GAN) is a deep-learning-based model introduced by Goodfellow~\citep{gan}, that allows to create synthetic data that is indistinguishable from real data. The model has two neural networks (a generator and a discriminator), and a training data set.\\
The generator's objective is to create data as similar as possible of the training data set. The discriminator's objective is to say which data comes from the generator and which data comes from the training data set just having the data as input. This way, the generator and discriminator competes against each other to obtain better results.\\
If the discriminator fails to tell the source of the data, it means the generator was able to fool the discriminator. In this case, the discriminator will be the one learning from its mistake. If the discriminator is able to tell correctly the source of the data, the generator is the one that will be learning. When the training is complete, the generator network is the one used to keep generating data.\\

\begin{figure}[H]
  \includegraphics[width=\linewidth]{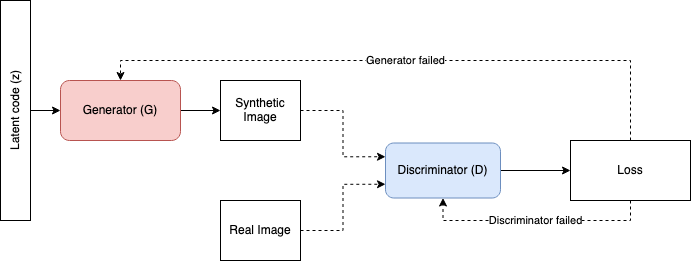}
  \caption{GAN architecture.}
  \label{fig:gan}
\end{figure}

In our case, the data are images of faces. In the training, the generator will be learning the structure of these images, projecting them to a $\mathbb{R}^{n}$ space called \textbf{latent space}. This way, the generator takes as an input a $\mathbb{R}^{n}$ point, usually called \textbf{latent code} or $z$, and the output is the image in that point of the latent space.

\subsubsection{Loss function}

The generator ($G$) and the discriminator ($D$) are competing against each other, playing a two-player minimax game. $G$ receives some input ($z$) and returns a synthetic image. $D$ receives an image, and tries to tell if it is a real image or a synthetic one. This competition can be described with the following function~\citep{gan}.

\begin{equation}
    \min_{G}\max_{D}\mathbb{E}_{x\sim p_{\text{data}}(x)}[\log{D(x)}] +  \mathbb{E}_{z\sim p_{\text{z}}(z)}[1 - \log{D(G(z))}]
\end{equation}

\subsubsection{Wasserstein GAN (WGAN)}

WGAN~\citep{wgan} is an alternative algorithm to the traditional GAN algorithm. It proposes a new loss function using Wasserstein distance, the original GAN loss function uses the Jensen-Shannon divergence instead. This change improves the stability of learning and solve problems like mode collapse that the traditional GAN has. 

\subsubsection{Deep Convolutional GAN (DCGAN)}

DCGANs~\citep{dcgan}, in comparison with the traditional GANs, changes the architecture of the Generator and Discriminator. This model uses convolutional networks for both of them. The most important result this model gives is the stability it has.

\subsubsection{Conditional GAN (cGAN)}

In the traditional GAN, there is no control in the outputs of the generator. This is why cGANs~\citep{cgan} are introduced. A cGAN is an extension of GAN that introduces the possibility of adding labels to the input in order to influence the generator and discriminator networks. This gives some control to the synthetic images the generator creates. It also presents the possibility of generating new tags to an image.

\subsubsection{Auxiliary Classifier GAN (AC-GAN)}

AC-GAN~\citep{acgan} is an extension of cGANs that modifies the discriminator to also predict the label (also known as class or auxiliary classifier) of the input image. 

\subsubsection{Information Maximizing GAN (InfoGAN)}
An InfoGAN~\citep{infogan} is an extension to GANs that is based in information theory. InfoGANs are able to learn disentangled representations in a completely unsupervised manner. This representations are competitive to the ones learned with surpervised methods.

\subsubsection{Pix2Pix}
Pix2Pix~\citep{pix2pix} is a software that uses a cGAN for image-to-image translation. It can be used for different purposes. Some of the known applications are labels-to-street-scene, aerial-to-map, labels-to-facade, day-to-night and edges-to-photo.

\begin{figure}[H]
  \includegraphics[width=\linewidth]{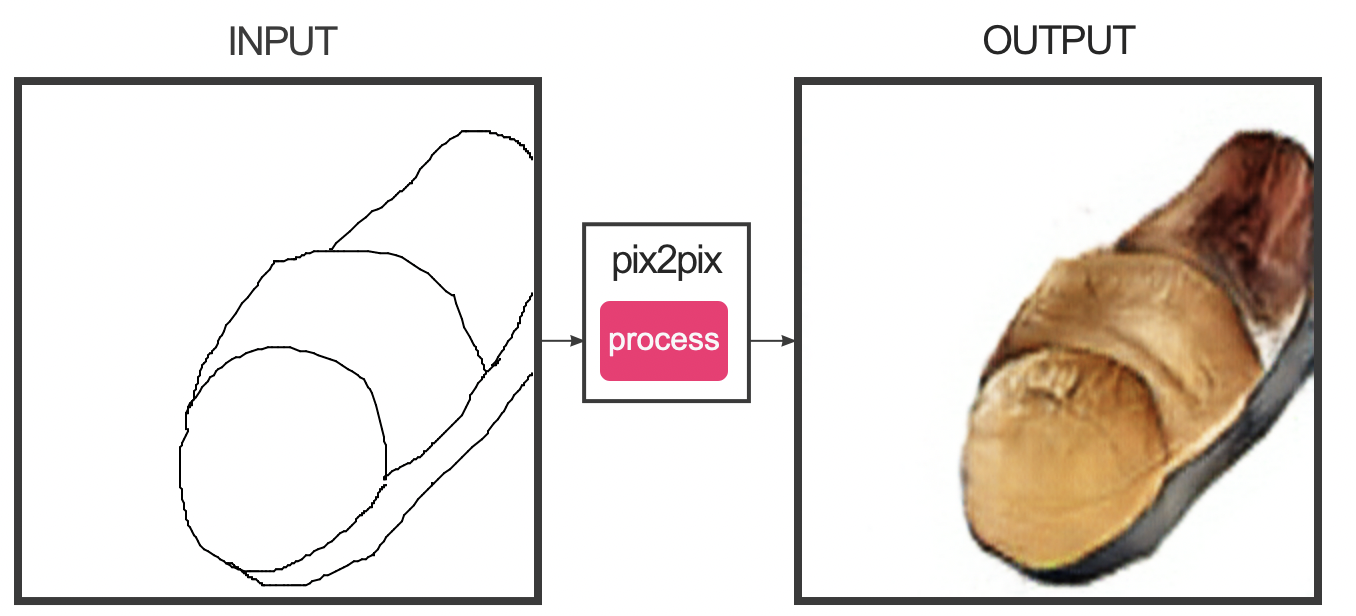}
  \caption{An example from a edges-to-shoe image translator that uses Pix2Pix.}
  \label{fig:pix2pix}
\end{figure}

\subsubsection{Stacked GAN (StackGAN)}
\begin{figure}[H]
  \includegraphics[width=\linewidth]{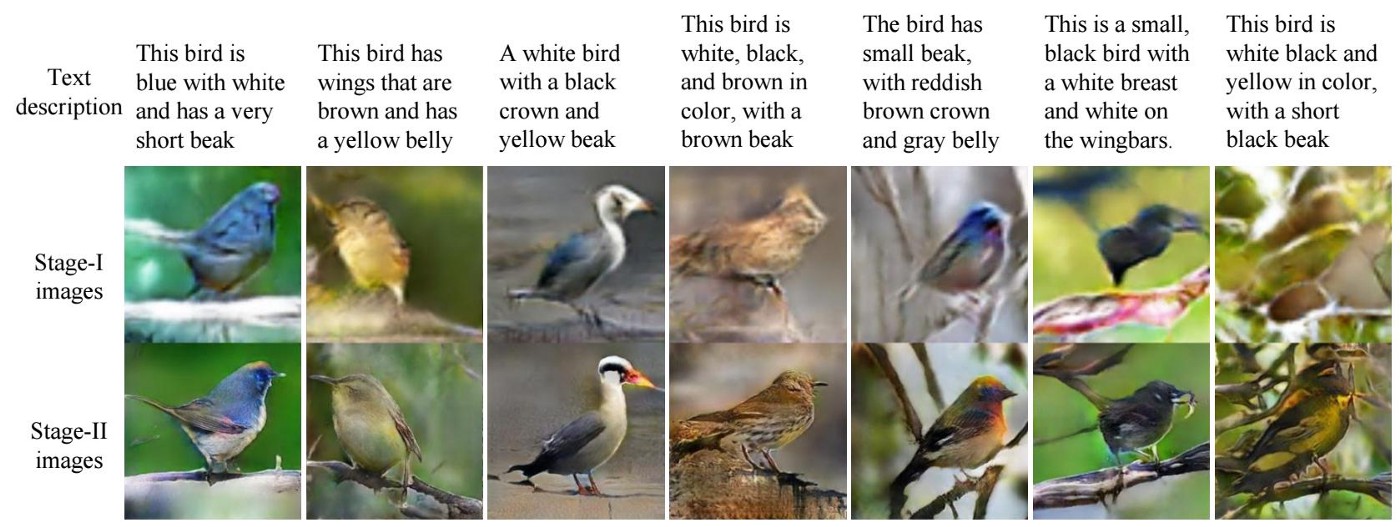}
  \caption{Results from StackGAN paper~\citep{stackgan}.}
  \label{fig:stackgan}
\end{figure}

StackGAN~\citep{stackgan} proposes an architecture that chains multiple GANs together to create photo-realistic images. The output of one of the GANs is the input of the next one. The idea is to give more resolution to the output image throughout each GAN it passes through. When introduced, it was tested with a text-to-image translation, giving some interesting results, as seen in figure \ref{fig:stackgan}.

\subsubsection{Cycle-Consistent GAN (CycleGAN)}
CycleGAN~\citep{cyclegan} is an extension to GAN that connects two GANs in a cycle. Its objective is to obtain two mapping functions that translates from one set of images to another. For example, some of the translations can be horses-to-zebras, artist-to-photo and winter-to-summer, and their respective inverses.

\subsubsection{Progressive Growing GAN (PGGAN)}
PGGAN~\citep{pggan} introduces a new training methodology for GANs. First it trains a GAN to generate 4x4 images, then it adds another layer to both the generator and discriminator to generate 8x8 images, and so on until the desired resolution (the paper goes up to 1024x1024). The idea behind the algorithm is to first learn greater-scale structure and later on focus in the fine details. This methodology reduces training time and obtains more realistic results.

\subsubsection{Style-Based GAN (StyleGAN)}

StyleGAN~\citep{stylegan2} is an extension of PGGAN that modifies the generator architecture. GANs used to use the vector $z$ directly, but StyleGAN maps it to a new vector $w$. This new vector $w$ is divided into different layers which are used together with Gaussian noise vectors as inputs of the different layers of the StyleGAN's generation algorithm.

\begin{figure}[H]
  \includegraphics[width=\linewidth]{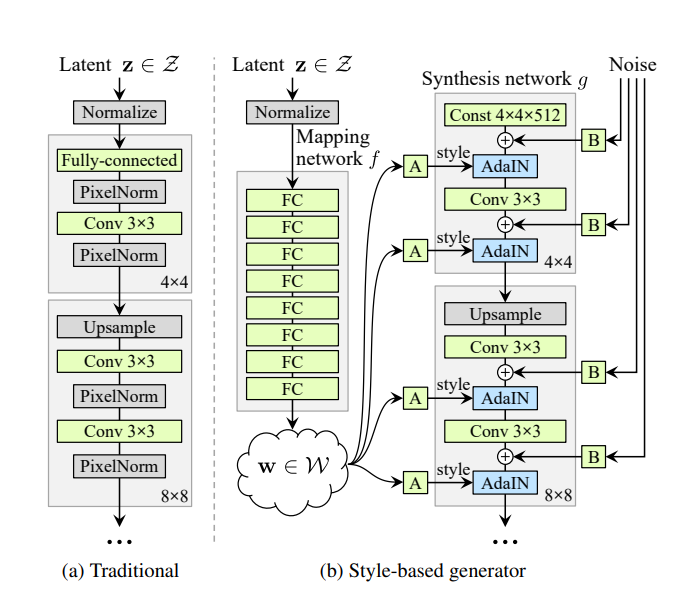}
  \caption{Traditional GAN architecture vs StyleGAN architecture.~\citep{stylegan2}}
  \label{fig:stylegan}
\end{figure}

\subsection{Facial properties modification within GAN}
In 2020, Erik Härkönen, Aaron Hertzmann, et al. released a paper titled \emph{GANSpace: Discovering Interpretable GAN Controls}~\citep{ganspace_directions} showing how to find latent directions allowing to modify the properties of the objects in the generated image by the neural network. This was mainly based on Principal Component Analysis (PCA), which provides a much faster way to extract meaningful latent directions in comparison to manual supervision or expensive optimization. The results of this paper can be applied to StyleGAN's 2 architecture, paving the way to modify facial attributes of a generated image within the neural network itself.

\subsection{Face editing with GAN}
Another paper was released in 2022 focusing in face editing with GANs moving across a direction in the latent space. \emph{Face editing with GAN’s – A Review}~\citep{face_editing_gan}, written by Parthak Mehta, Sarthak Mishra, et al, shows how can other classification models can be used, such as logistic regression or SVM, to find the directions that represent a feature in the trained StyleGAN model. Not only that, but they also found a way to find a latent vector which generated image is similar to a target one. This way, a real portrait can be used as an input to find a similar face generated by the neural network, which attributes could then be modified using the directions that represent a feature.

\subsection{Initial project status}
This project aims to continue the work of Jimena Lozano and Maite Herrán Oyhanarte on the subject of GANs to create and manipulate facial images~\citep{lozanoherran}. This tool was intended as a software to generate realistic facial images with the ability to manipulate their main characteristics, such as eye size and separation, nose size, etc. The resulting software is able to generate new random, artificial facial images from scratch and generate transitions between two generated pictures. StyleGAN 2 neural network was the core of this software to provide those features, paired with an API and a Web application built in Python and React respectively. All of these made possible an initial version of a software that enabled the Lab to improve their preparation workflow for their experiments.

\biblio % Needed for referencing to working when compiling individual subfiles - Do not remove

%% file: 01_Chapters/03FaceFrame.tex
\label{section:face_frame_materials}

\subsection{Face-frame variation measurement}
\begin{figure}[h]
  \includegraphics[width=\linewidth]{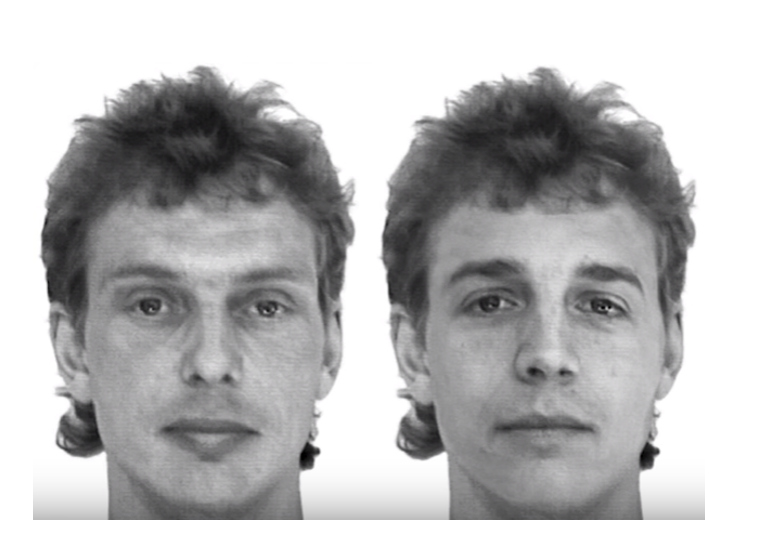}
  \caption{Two different faces with the same face-frame.}
  \label{fig:faceframe}
\end{figure}

There is no standardized way of defining a face-frame. We define it as the set of characteristics of a face that gives context for the internal characteristics of such face. A face-frame includes the contour of the face, hair, ears. A face-frame does not include eyes, mouth, nose or eyebrows. For example, Figure \ref{fig:faceframe} shows two faces with the same frame. The hypothesis of the Lab is that this two faces may be mistaken with each other because they share the same frame.

Having this in mind, we measure the variation of the face-frame between two images based on the position of the face and hair in the image.

\subsubsection{Image segmentation}
\begin{figure}[h]
  \includegraphics[width=\linewidth]{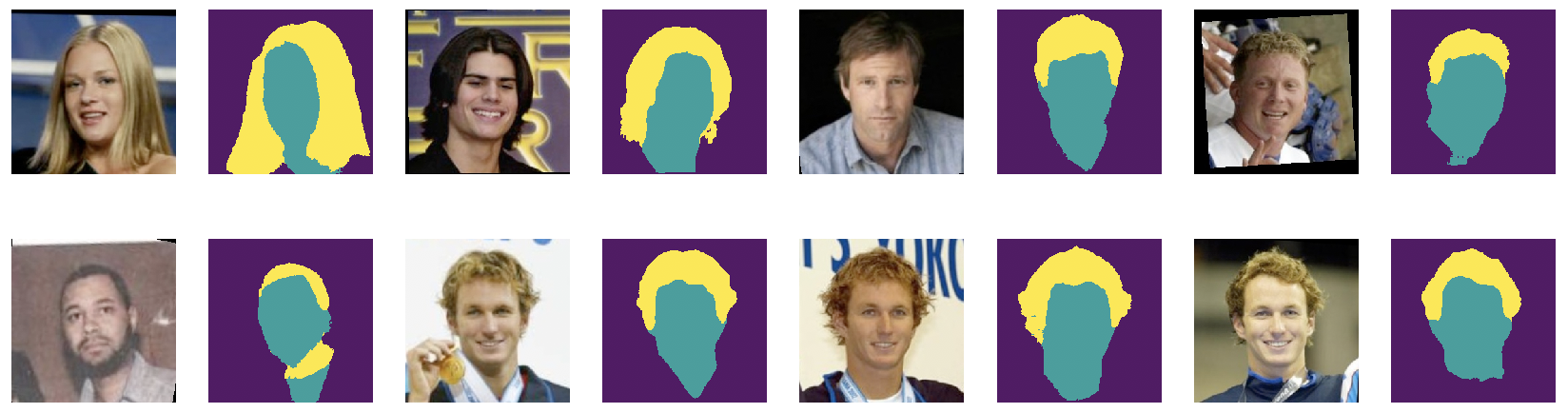}
  \caption{Segmentation made by the neural network.}
  \label{fig:segmentation}
\end{figure}

The first thing we need to do is to identify the face and hair inside an arbitrary image. This consists of a simple segmentation problem, separate the pixels of an image in three groups: face, hair and background. For this, an open-source pretrained neural network is used~\citep{segmentationgithub}. This network proved to give satisfying results for different faces and images with different conditions. This is essential for the measurement because there is no need to modify any parameters for different images. It is worth mentioning that although the neck is not part of the face-frame, the Lab thinks is better if it is preserved. Therefore, for the objective of this work, it is acceptable that the neural network classifies the neck as part of the face. In Figure \ref{fig:segmentation}, we can see some examples of segmentation made by this network. In yellow the pixels classified as hair; in blue the pixels classified as face; and in purple the pixels classified as background.

\subsubsection{Face-frame variation formula}
\label{section:function}
Once each pixel of each of the two images is classified, we compare the classifications between the pixels in the same position of the two images. Let $f$ be a function that compares the classifications of two pixels,
\begin{equation}
f(c_1, c_2) = 
   \left\{
\begin{array}{ll}
      0 & c_1 = c_2 \\
      0.2 & c_1 = "Face" \land c_2 = "Hair" \\
      0.2 & c_1 = "Hair" \land c_2 = "Face" \\
      1 & otherwise \\
\end{array} 
\right. 
\end{equation}

The number 0.2 is arbitrary, but it means that the face-hair variation is 5 times less important than variations involving the background of the image.\\

Let F be the function that measures the variation of face-frame between two images ($I_1$ and $I_2$) of same size, with height $h$ and width $w$. The classifications of the pixels of $I_1$ are $p_{i,j}$, and the classifications of the pixels of $I_2$ are $q_{i,j}$.

\begin{equation}
F(I_1, I_2) = \frac{\sum_{i=1}^{h} \sum_{j=1}^{w} f(p_{i,j}, q_{i,j})}{h*w}
\end{equation}

This function yields a percentage of the images that is classified differently, giving more importance to background variations. If the face-frame does not vary between two images $I_1$ and $I_2$, then $F(I_1, I_2)=0$. If there is a change in at least one pixel, then $0 < F(I_1, I_2) \leq 1$.

\label{section:correction}
\subsection{Correction for image projection to latent space}
\begin{figure}[h]
  \includegraphics[width=\linewidth]{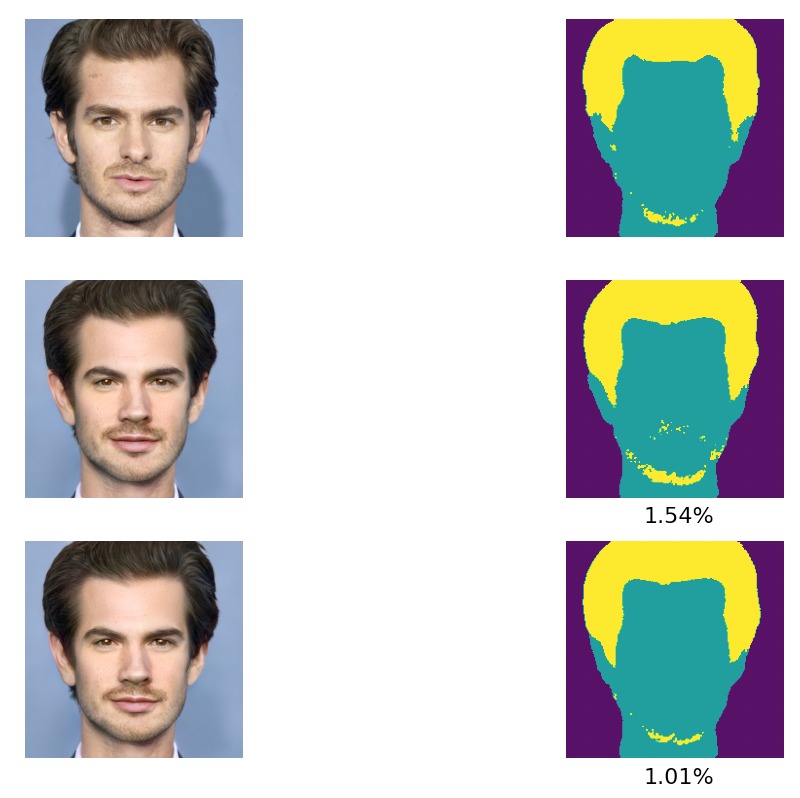}
  \caption{A target face image, its output from the neural network and its output after running it through the correction algorithm, with its face-frame deviation.}
  \label{fig:correction_demo}
\end{figure}

As mentioned in section \ref{section:projection}, StyleGAN2 uses a metric~\citep{zhang} to estimate perceptual similarity to the target image. Albeit this method works pretty well, it does not take into account the face-frame variation. To try to obtain results with the least variation of face-frame as possible, a post-processing algorithm (henceforth referred to as \textbf{face-frame correction}) has been designed to pick the best image, which has the lower variation of them all, using the function defined in \ref{section:function} to measure this metric.\\
The correction takes the target image and the latent code of the projection as an input. First, it calculates the standard deviation of 10000 random latent codes that create realistic images, which allows us to apply noise to the original latent code in a way that we can be certain it will yield a realistic image. The $f(target image, G(initial latent code))$ is also calculated and stored in this first step. The noise strength in each iteration is the result of applying the following formula:
\begin{equation}
strength(i) = l_{std} * noise_0 * (\frac{1-\frac{i}{10000}}{nrl}) ^{2}
\end{equation}
Being $i$ the current iteration number, $l_{std}$ the standard deviation of the latent codes previously mentioned, $noise_0$ a constant representing the initial noise factor, and $nrl$ a constant representing the noise ramp length. In these experiments, the values are $noise_0 = 0.005$ and $nrl = 0.75$, which follow what StyleGAN2 uses \citep{stylegangithub}\\
Throughout every of the $n$ iterations, the algorithm introduces a Gaussian noise multiplied by $strength(i)$ to the latent code. Depending on the constants $noise_0$ and $nrl$, it can be expected that the output image differs too little from the original one so that, if this new latent code generates an image with less face-frame variation than the previous latent code, then this new one is stored and used as the current latent code. If that is not the case, the original is preserved. In Figure \ref{fig:correction_demo}, we can see an example of a target face being projected into the latent space as it outputs the neural network and another one after applying the correction algorithm.

Having set fixed values of $noise_0 = 0.005$ and $nrl = 0.75$ taken from StyleGAN2 \citep{stylegangithub}, the optimal value of $n$ is going to be studied having in mind the reduction of the face-frame variation of the target image and the projected one as much as possible.

\biblio % Needed for referencing to working when compiling individual subfiles - Do not remove

%% file: 01_Chapters/04MaterialsAndMethods.tex
\label{section:materials}

\subsection{Projecting an arbitrary image to latent space}
\label{section:projection}
The image to latent space projection operation StyleGAN2 provides, takes advantage of the fact that the latent space is semantically smooth. This means that small changes in the input vector, will result to small changes in the resulting image. A random input vector is used to start with. This vector is slightly modified and the resulting image is compared to the target image. It uses a standard LPIPS metric to estimate perceptual similarity between an image and the target image~\citep{zhang}. If the modification resulted in a better similarity to the target image, it is maintained. This process of slight modifications and measurements is done a thousand times.\\
The target image's frame and the resulting image's frame are compared.\\
\begin{figure}[H]
  \includegraphics[width=\linewidth]{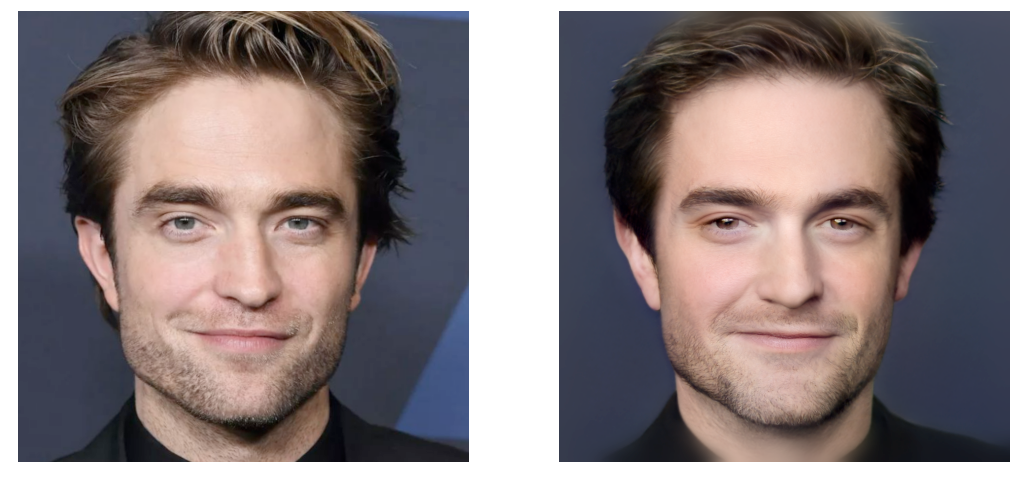}
  \caption{To the left, the target image. To the right, its projection to the latent space}
  \label{fig:projection}
\end{figure}

\subsection{Operations on the target image to measure face-frame variation}
We can take advantage of the fact that there are multiple latent directions we can move through, modifying certain attributes of an image. Some of this latent directions are of interest as they can be used by the Lab to edit an image right in the neural network itself. We define an operation done on an image which originates by a latent code as an addition or subtraction of a vector in such a way that a property of the face is modified.\\
Having this in mind, we will study the face-frame variation when applying some of these operations. Let $O$ be the operation to be studied. The face-frames to compare are the ones from $I_1$ and $O(I_1)=I_2$. As the results may vary from image to image, the measurement is taken as the average measurement of a hundred different images. The same hundred images are used in each operation.

\subsubsection{Moving through latent directions}

\paragraph{Age}
The age direction transforms the face in a way that it looks like older or younger version of the original face.
\begin{figure}[H]
  \includegraphics[width=\linewidth]{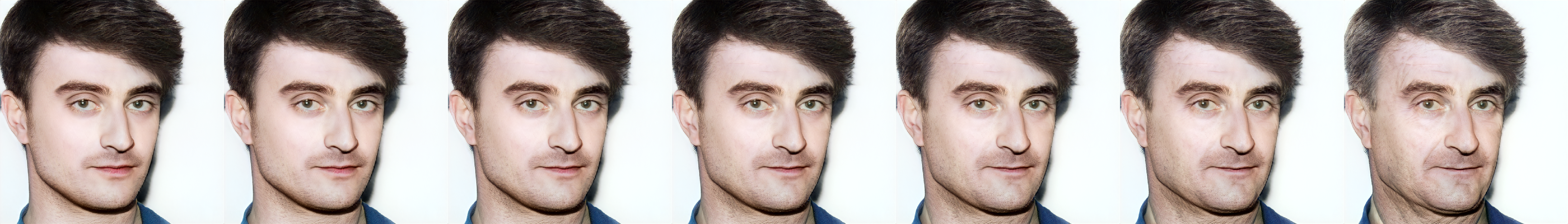}
  \caption{Images generated moving through age direction}
  \label{fig:age}
\end{figure}

\paragraph{Gender}
The gender direction transforms the face in a way that it looks like a more masculine or feminine version of the original face.
\begin{figure}[H]
  \includegraphics[width=\linewidth]{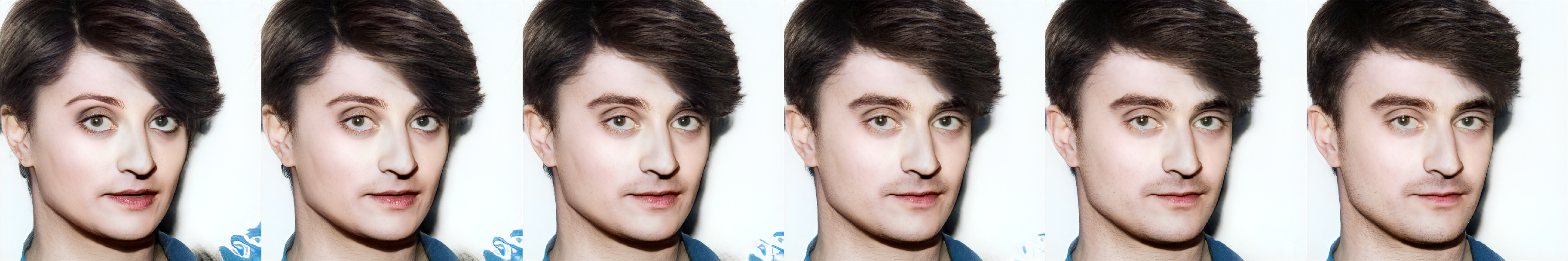}
  \caption{Images generated moving through gender direction}
  \label{fig:gender}
\end{figure}

\paragraph{Horizontal orientation}
The horizontal orientation direction transforms the face in a way that it looks like rotating horizontally.
\begin{figure}[H]
  \includegraphics[width=\linewidth]{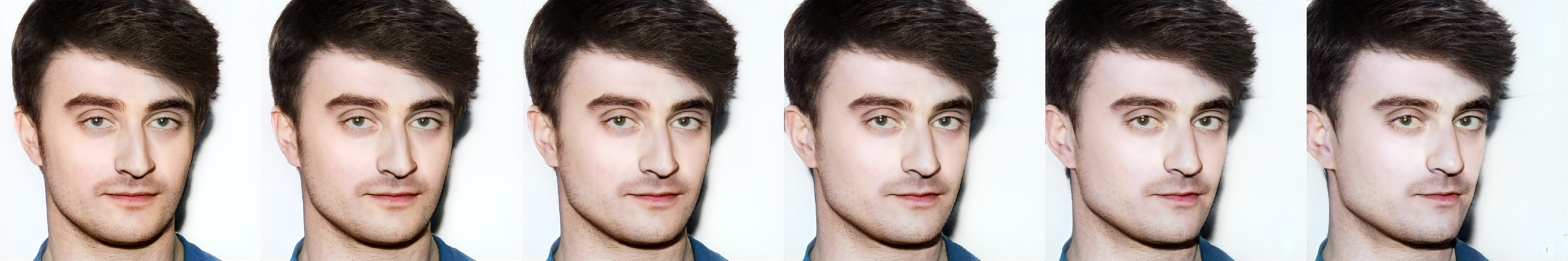}
  \caption{Images generated moving through horizontal face orientation direction}
  \label{fig:horizontal}
\end{figure}

\paragraph{Vertical orientation}
The vertical orientation direction transforms the face in a way that it looks like rotating vertically.
\begin{figure}[H]
  \includegraphics[width=\linewidth]{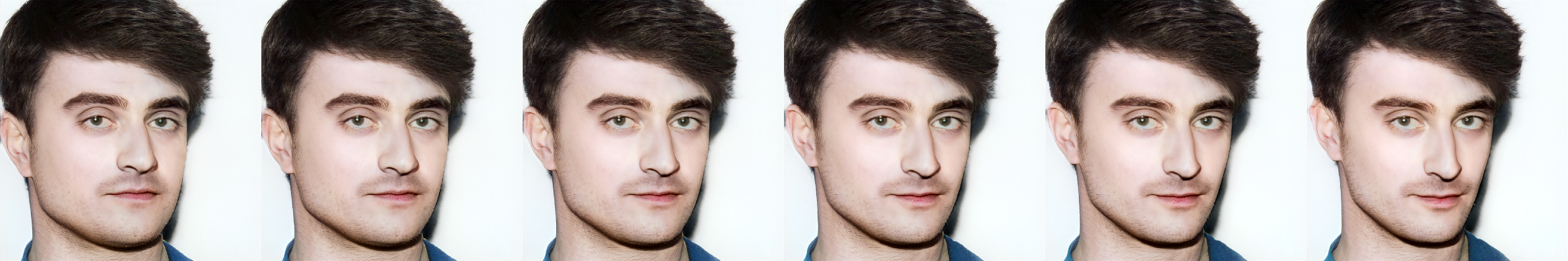}
  \caption{Images generated moving through vertical face orientation direction}
  \label{fig:vertical}
\end{figure}

\paragraph{Eyes open}
The eyes open direction transforms the face opening or closing its eyes.
\begin{figure}[H]
  \includegraphics[width=\linewidth]{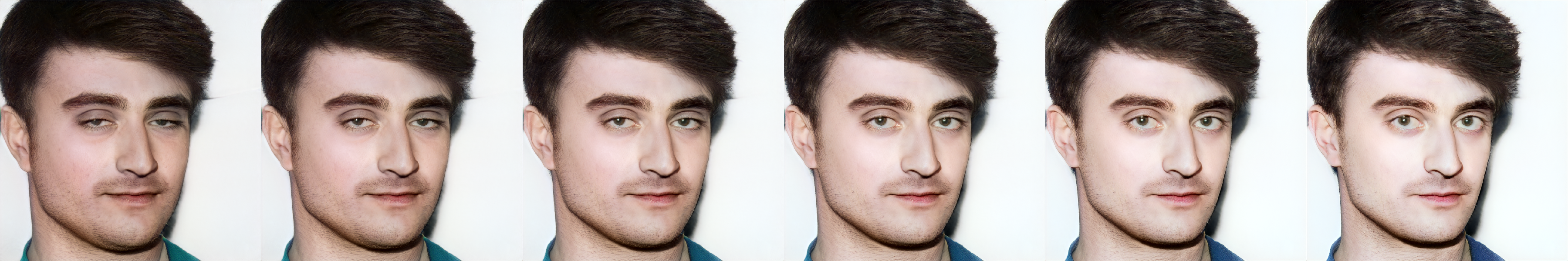}
  \caption{Images generated moving through eyes open direction}
  \label{fig:eyesopen}
\end{figure}

\paragraph{Mouth open}
The mouth open direction transforms the face opening or closing its mouth.
\begin{figure}[H]
  \includegraphics[width=\linewidth]{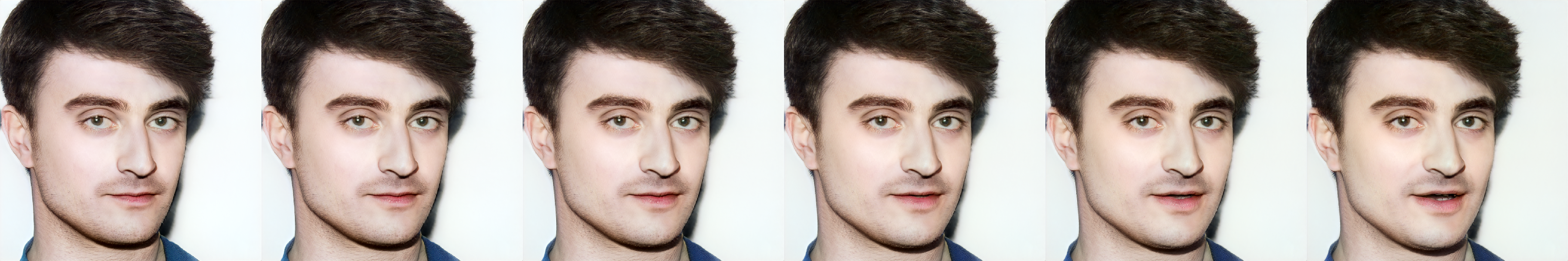}
  \caption{Images generated moving through mouth open direction}
  \label{fig:mouthopen}
\end{figure}

\paragraph{Smile}
The smile direction transforms the face adding or removing a smile.
\begin{figure}[H]
  \includegraphics[width=\linewidth]{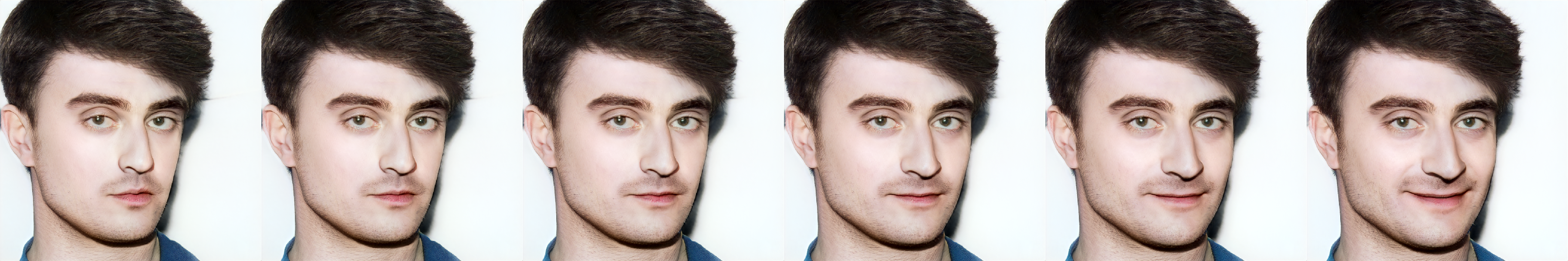}
  \caption{Images generated moving through smile direction}
  \label{fig:smile}
\end{figure}

\subsubsection{Style mixing}
Style mixing is an operation that takes two images, and generate other two images exchanging their styles.~\citep{stylegan2} This means that their layout of the face is maintained but the style of the image will be the one of the other input image.
\begin{figure}[H]
  \hfill\includegraphics[width=0.5\linewidth]{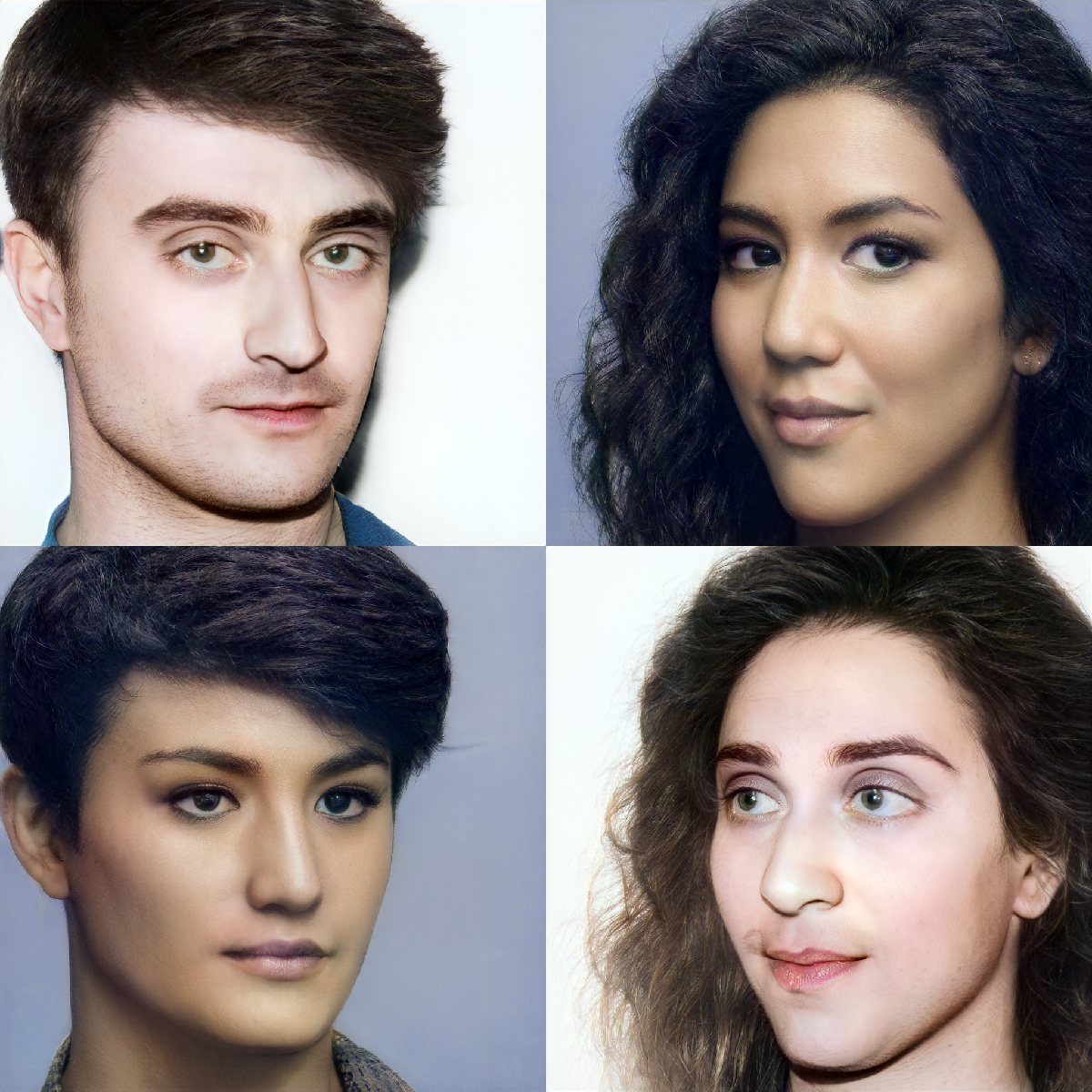}\hspace*{\fill}
  \caption{On the top row, the two input images. On the bottom row, the two images with their styles mixed}
  \label{fig:stylemix}
\end{figure}

\biblio % Needed for referencing to working when compiling individual subfiles - Do not remove

%% file: 01_Chapters/05Experiments.tex
\label{section:experiments}

\subsection{Projection to latent space}
\label{section:experiments_projection}
In order to be examine the effectiveness of the face-frame correction algorithm proposed in Section \ref{section:projection}, we first need to take a look at how well StyleGAN2 does when projecting a target image to the latent space in terms of face-frame variation. To measure this, we project 10 images of different faces to the latent space without using the correction algorithm, and we measure the variation using the function defined in Section \ref{section:function}. These results are then processed into a table allowing us to have a baseline of how much the variation is using the default StyleGAN2's function.

\subsection{Correction for projection}

After examining the variation that StyleGAN2 introduces, we can then study how well our proposed algorithm corrects this variation. In order to test this, we set $n = 2000$ and we run our algorithm for each of the 10 output images of the previous Section \ref{section:experiments_projection}. We then compare the variation in each iteration,  allowing us to study the optimal $n$ in terms of computational time and reduced variation.

\subsection{Latent directions}

We also want to take a look at the face-frame variation for the 10 images when transforming those moving across the aforementioned latent directions. This allows us to rank each operation in respect of the variation they introduce on the generated image, making them more or less elegible for image modification if face-frame stabilization is of the essence.

\biblio % Needed for referencing to working when compiling individual subfiles - Do not remove

%% file: 01_Chapters/06Results.tex
\label{section:results}

\subsection{Projection to latent space}

Table \ref{table:projection} shows the face-frame variation that the image to latent space projection achieved with each image. The maximum variation is at 3.192\% while the minimum at 0.747\%. The mean variation between these images is of 1.812\%. This results suggest that images are a decent starting point when trying to generate similar faces with minimum face-frame variation. 

\begin{table}[H]
\centering
\begin{tabular}{|l|l|}
\hline
\textbf{Image} & \textbf{Variation} \\ \hline
Andrew         & 1.429\%            \\ \hline
Daniel         & 1.894\%            \\ \hline
Emma           & 2.531\%            \\ \hline
Jennifer       & 1.790\%            \\ \hline
Kristen        & 2.267\%            \\ \hline
Matt           & 1.452\%            \\ \hline
Paul           & 1.198\%            \\ \hline
Rob            & 2.381\%            \\ \hline
Rosa           & 3.192\%            \\ \hline
Sheldon        & 1.041\%            \\ \hline
Zendaya        & 0.747\%            \\ \hline
\end{tabular}

  \caption{Face-frame variations results for projecting to latent space}
  \label{table:projection}
\end{table}

\begin{figure}[H]
  \includegraphics[width=\linewidth]{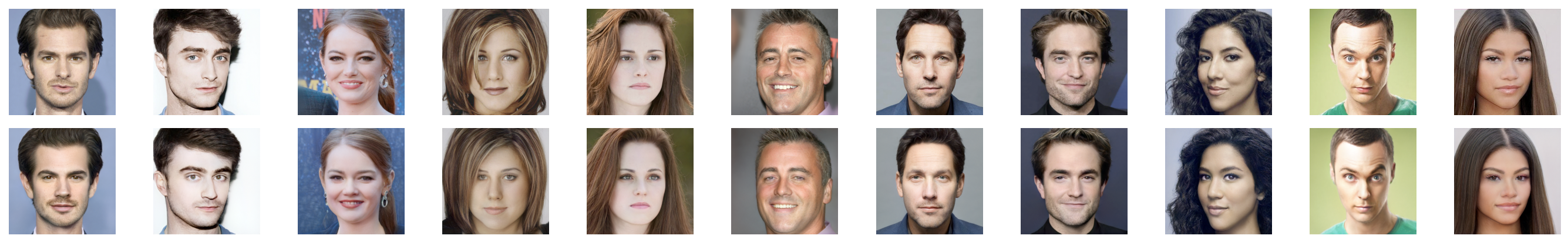}
  \caption{The 11 chosen faces and their respective projection generated by StyleGAN2 to their bottom.}
  \label{fig:projections}
\end{figure}

\subsection{Correction for projection}

\begin{figure}[H]
  \includegraphics[width=\linewidth]{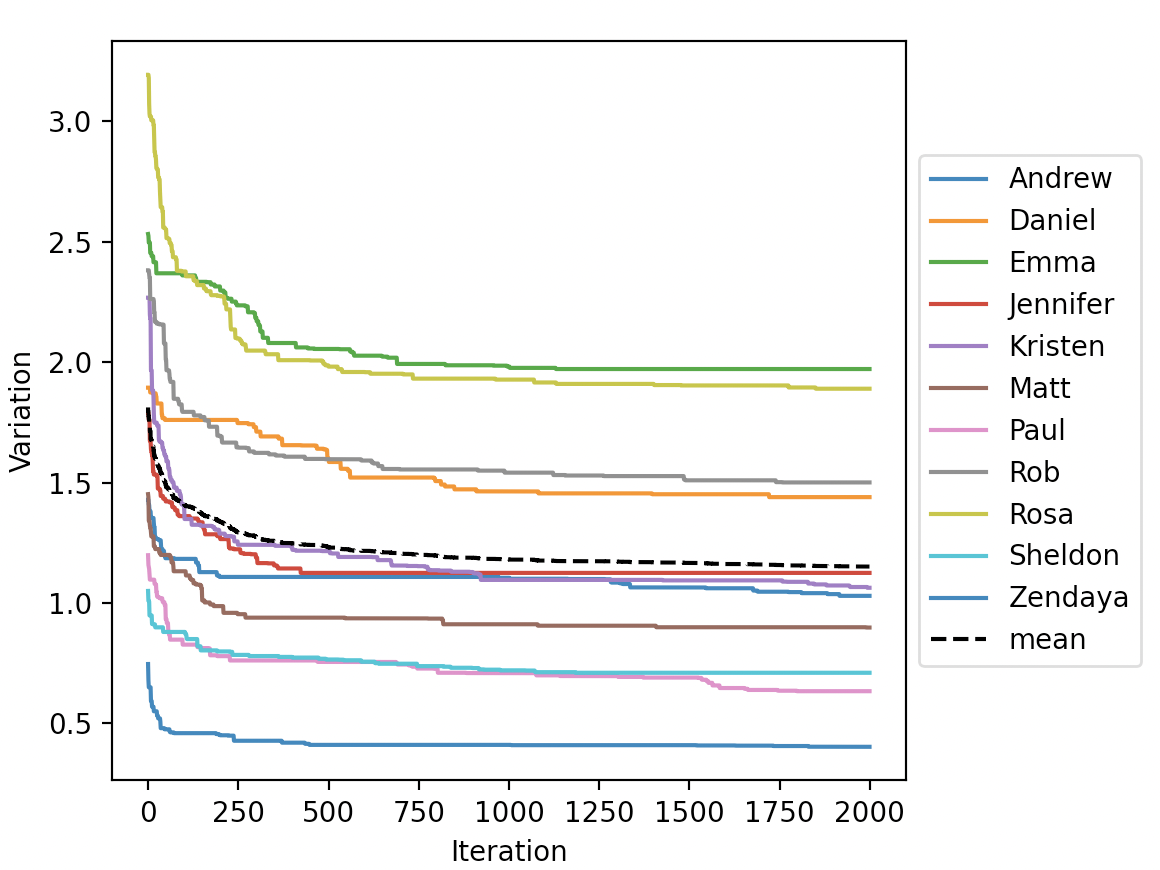}
  \caption{Graph of face-frame variations through the iterations}
  \label{fig:absolute}
\end{figure}

\begin{figure}[H]
  \includegraphics[width=\linewidth]{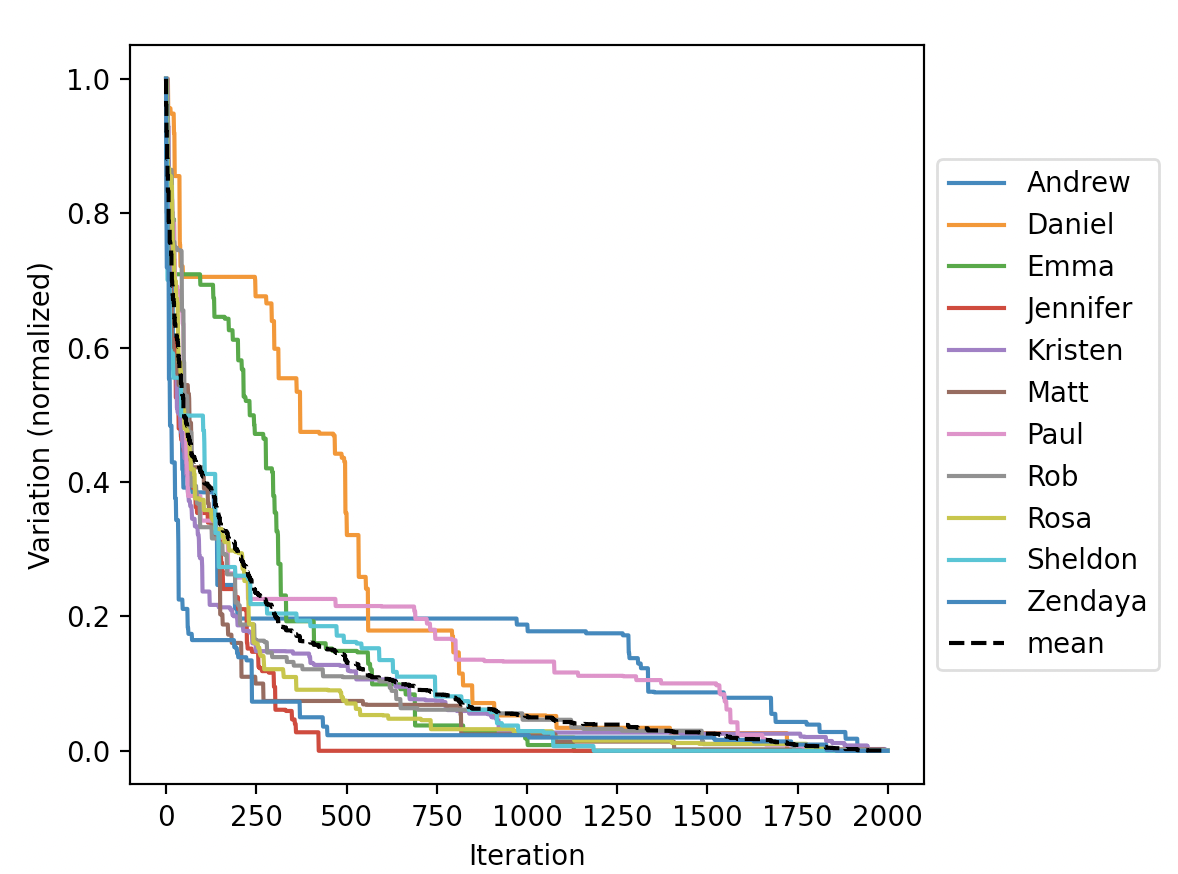}
  \caption{Graph of the normalized face-frame variations through the iterations}
  \label{fig:normalized}
\end{figure}

Figures \ref{fig:absolute} and \ref{fig:normalized} illustrate the improvement in the face-frame variation through the iterations of the correction. The curves of variation significantly decrease in the first 750 iterations. The last 1250 iterations takes account for the last 10\% of the improvement. It is notable that the improvement in the variation is always superior to the 20\% of the initial value, and in some of the cases, it goes over the 50\%. 

\begin{figure}[h]
  \includegraphics[width=\linewidth]{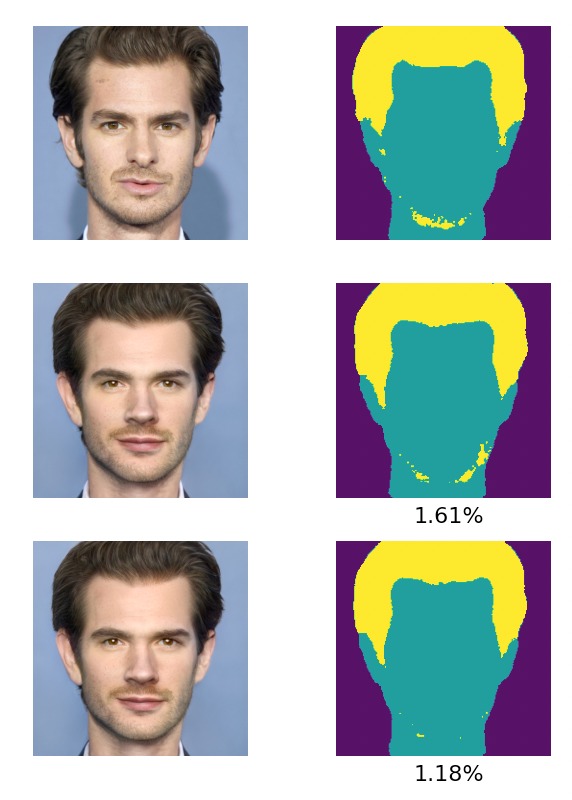}
  \caption{A target face image, its output from the neural network and its output after running it through the correction algorithm with 750 iterations, with its face-frame deviation.}
  \label{fig:750}
\end{figure}

\subsection{Latent directions}

\begin{figure}[H]
  \includegraphics[width=\linewidth]{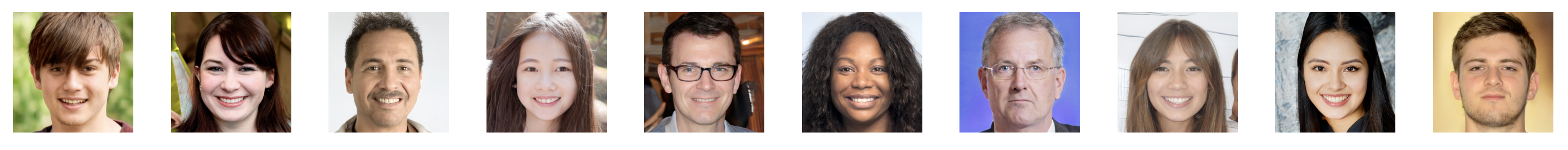}
  \caption{The 10 chosen faces for the latent directions experiments.}
  \label{fig:samples}
\end{figure}

We also check the face-frame variation for the 10 images when transforming those moving across the aforementioned latent directions. In Figure ~\ref{fig:samples}, we can see the chosen images. They are labeled as 1 to 10 from left to right.

When an image moves a great magnitude through a direction, they tend to give unexpected results, deforming the base structure of the face and giving unrealistic images. This is the reason the range measured in each direction changes in each case. For the set of chosen images, we decided the following ranges:

\begin{table}[H]
\centering
\begin{tabular}{|l|l|}
\hline
\textbf{Direction} & \textbf{Range} \\ \hline
Age                & [-3, 3]        \\ \hline
Gender             & [-3, 3]        \\ \hline
Vertical           & [-3, 3]        \\ \hline
Horizontal         & [-3, 3]        \\ \hline
Eyes-open          & [-10, 10]      \\ \hline
Mouth-open         & [-10, 10]      \\ \hline
Smile              & [-2, 2]        \\ \hline
\end{tabular}
    \caption{Ranges to measure for each direction}
  \label{table:ranges}
\end{table}

\begin{figure}[H]
  \includegraphics[width=\linewidth]{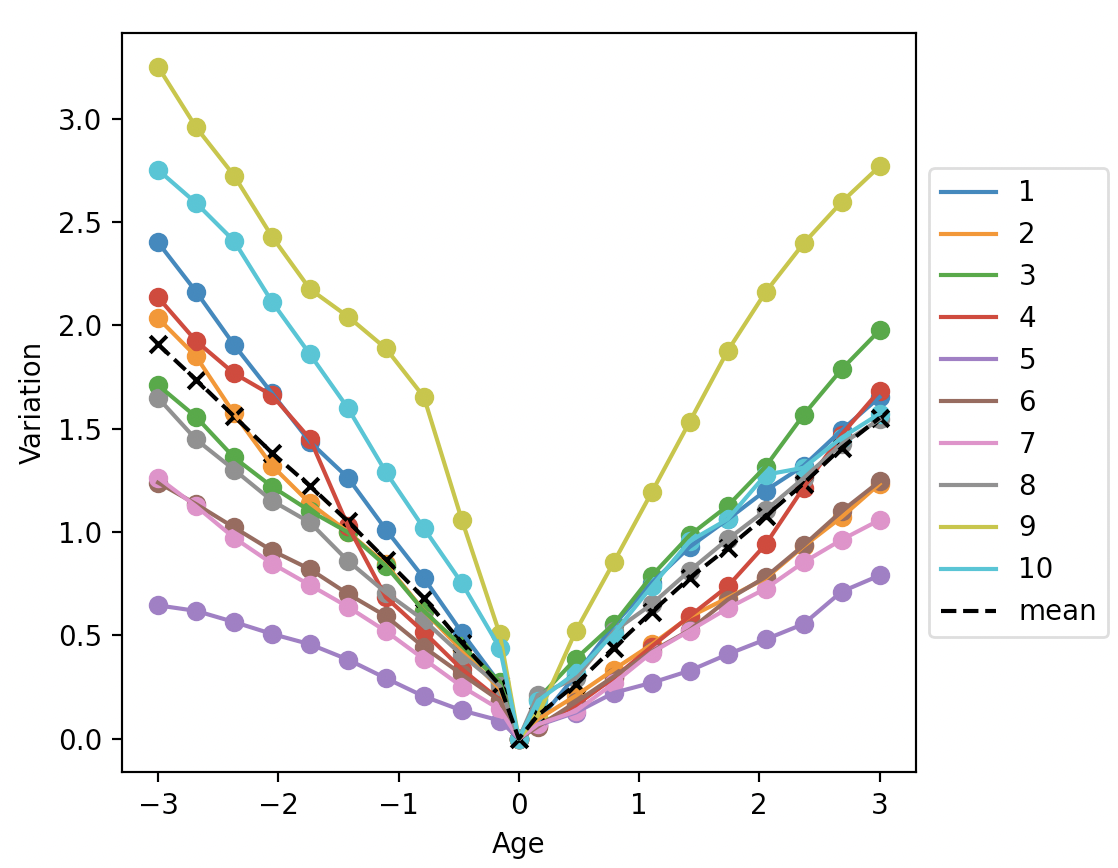}
  \caption{Graph of the face-frame variations through the age direction}
  \label{fig:graph_age}
\end{figure}

\begin{figure}[H]
  \includegraphics[width=\linewidth]{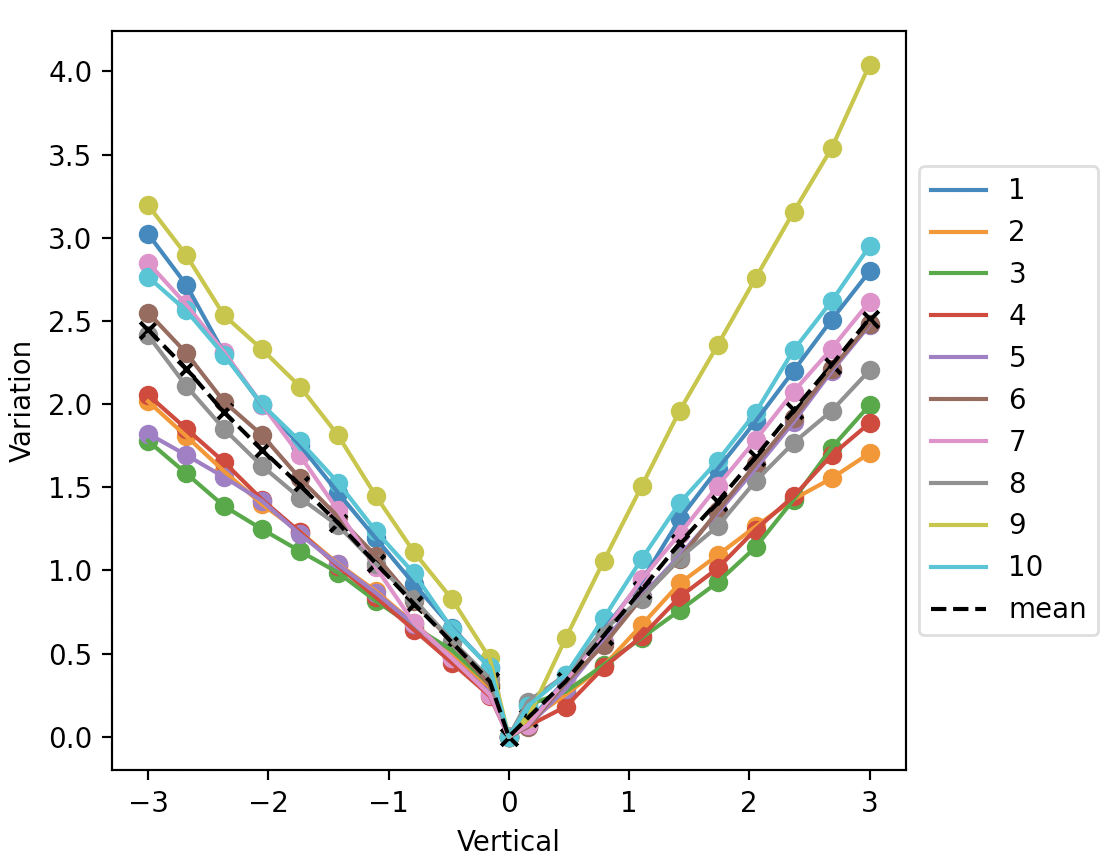}
  \caption{Graph of the face-frame variations through the vertical direction}
  \label{fig:graph_vertical}
\end{figure}

\begin{figure}[H]
  \includegraphics[width=\linewidth]{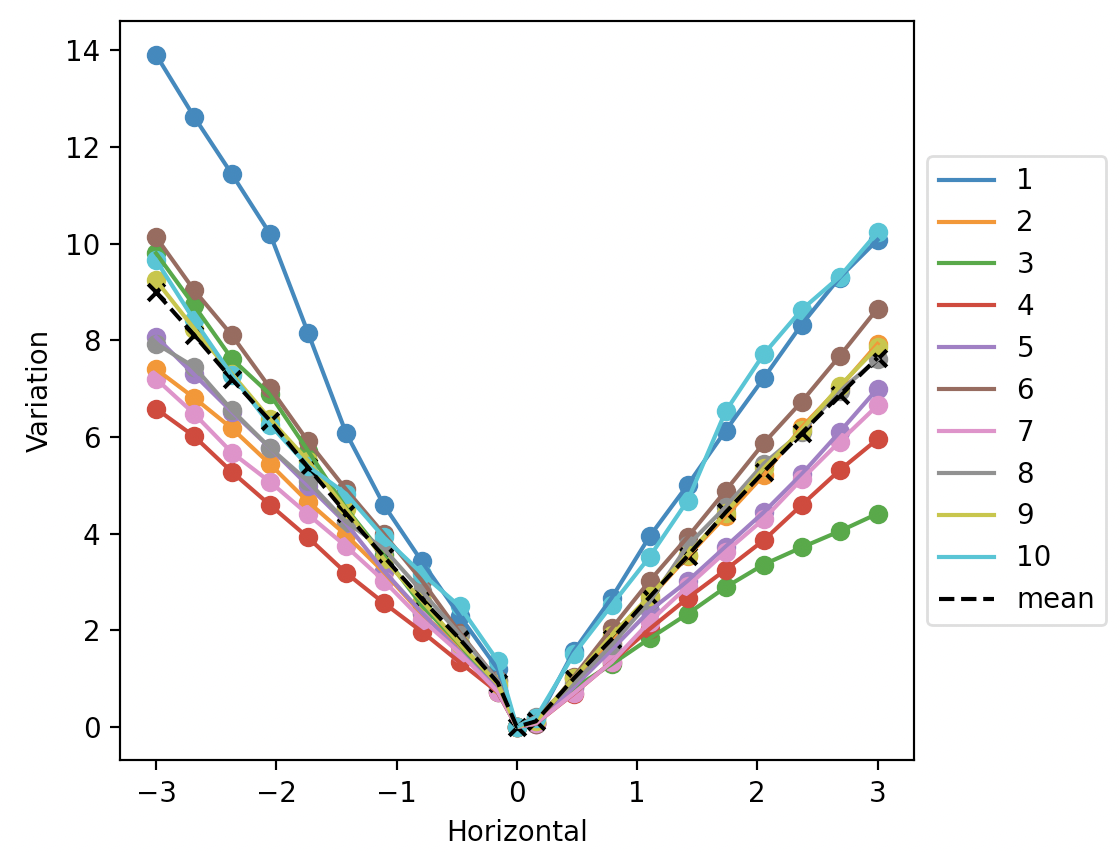}
  \caption{Graph of the face-frame variations through the horizontal direction}
  \label{fig:graph_horizontal}
\end{figure}

\begin{figure}[H]
  \includegraphics[width=\linewidth]{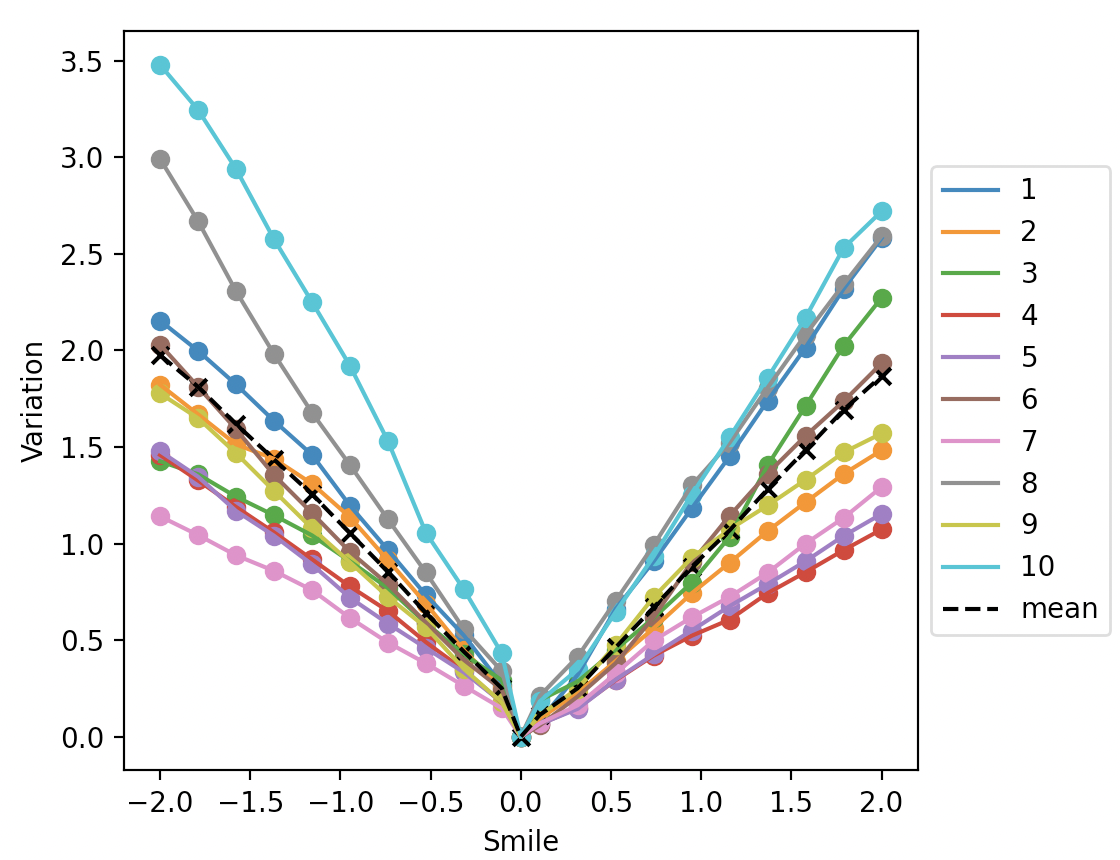}
  \caption{Graph of the face-frame variations through the smile direction}
  \label{fig:graph_smile}
\end{figure}

\begin{figure}[H]
  \includegraphics[width=\linewidth]{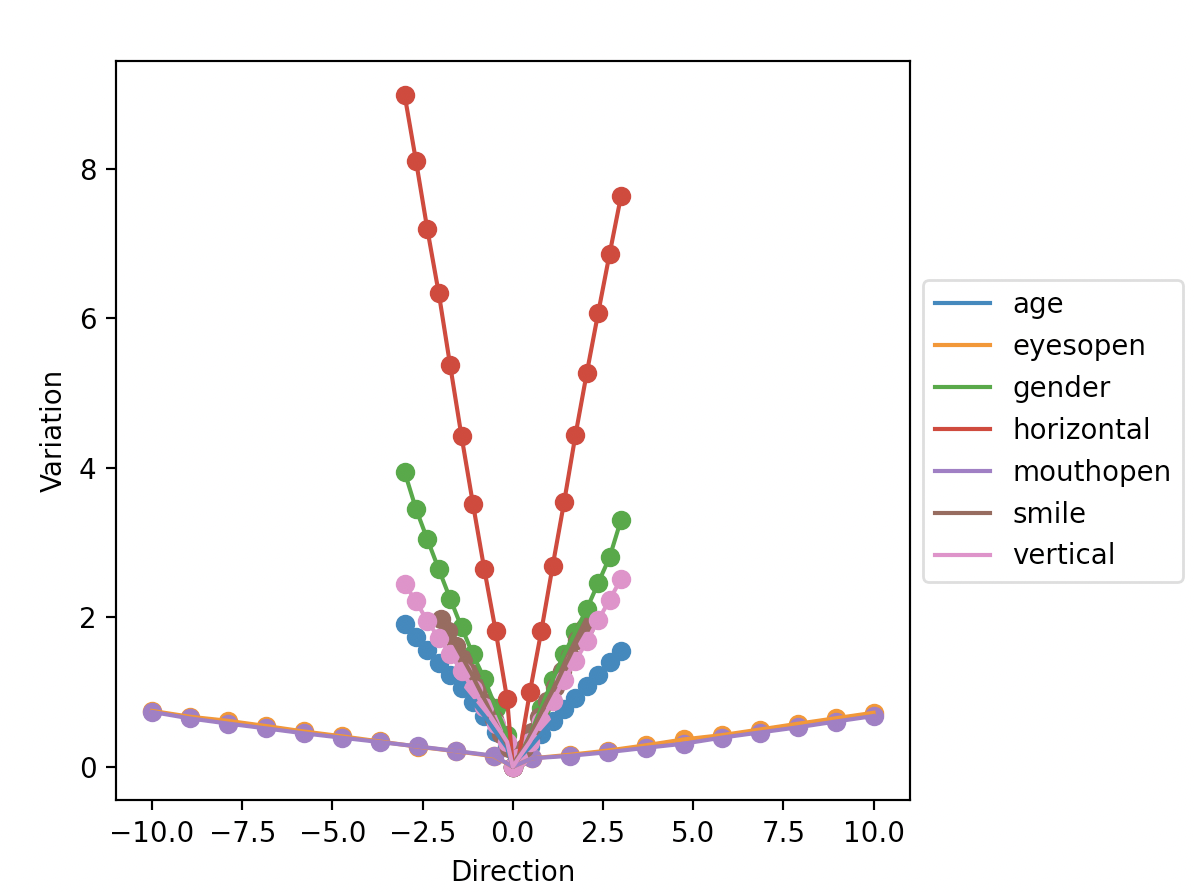}
  \caption{Graph of the mean face-frame variations through the each studied direction}
  \label{fig:graph_means}
\end{figure}

Figures \ref{fig:graph_vertical} and \ref{fig:graph_horizontal} both show that modifying the horizontal and vertical face perspective in the image greatly worsens the face-frame variation. Figure \ref{fig:graph_age} shows that gender modification is another operation that significantly increases the face-frame variation. These operations are some of the worst ones when considering this variation as can be seen in \ref{fig:graph_means}, together with smile modification (\ref{fig:graph_smile}) and age modification (\ref{fig:graph_age}).

\begin{figure}[H]
  \includegraphics[width=\linewidth]{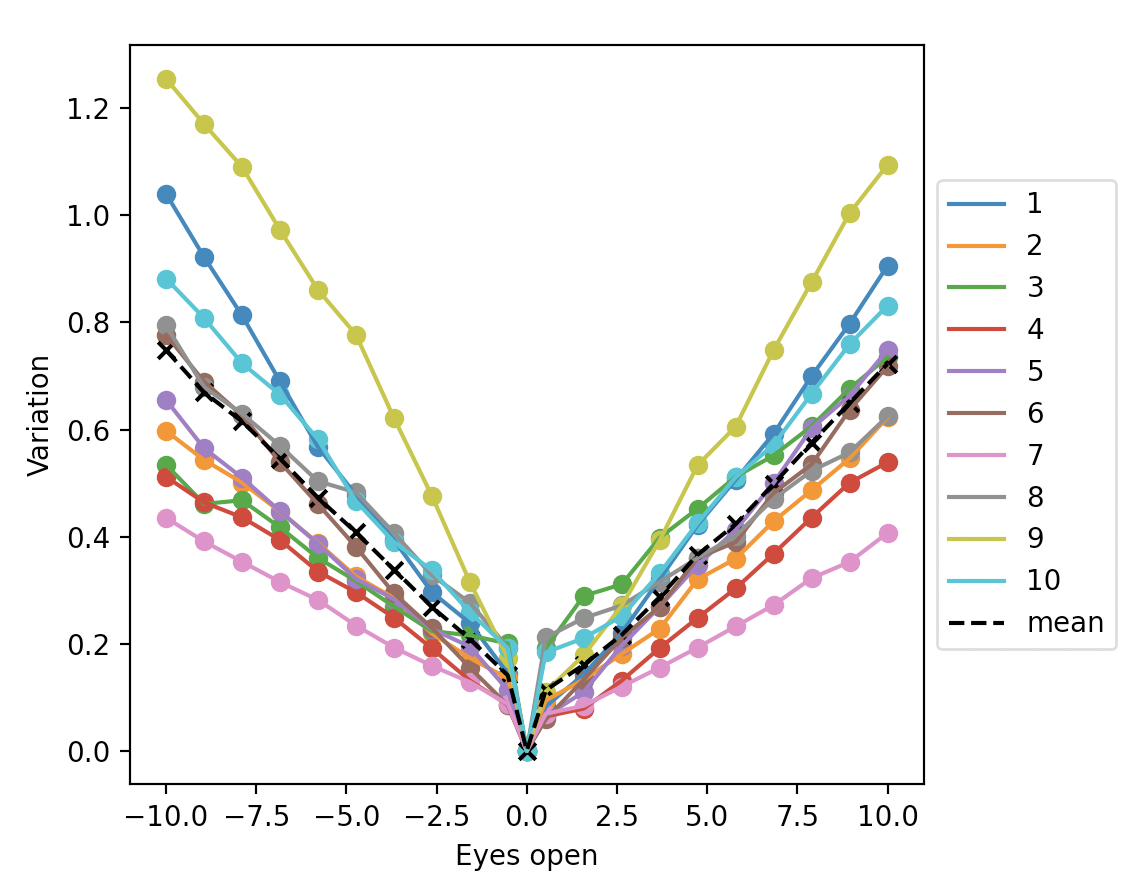}
  \caption{Graph of the face-frame variations through the eyes open direction}
  \label{fig:graph_eyesopen}
\end{figure}

\begin{figure}[H]
  \includegraphics[width=\linewidth]{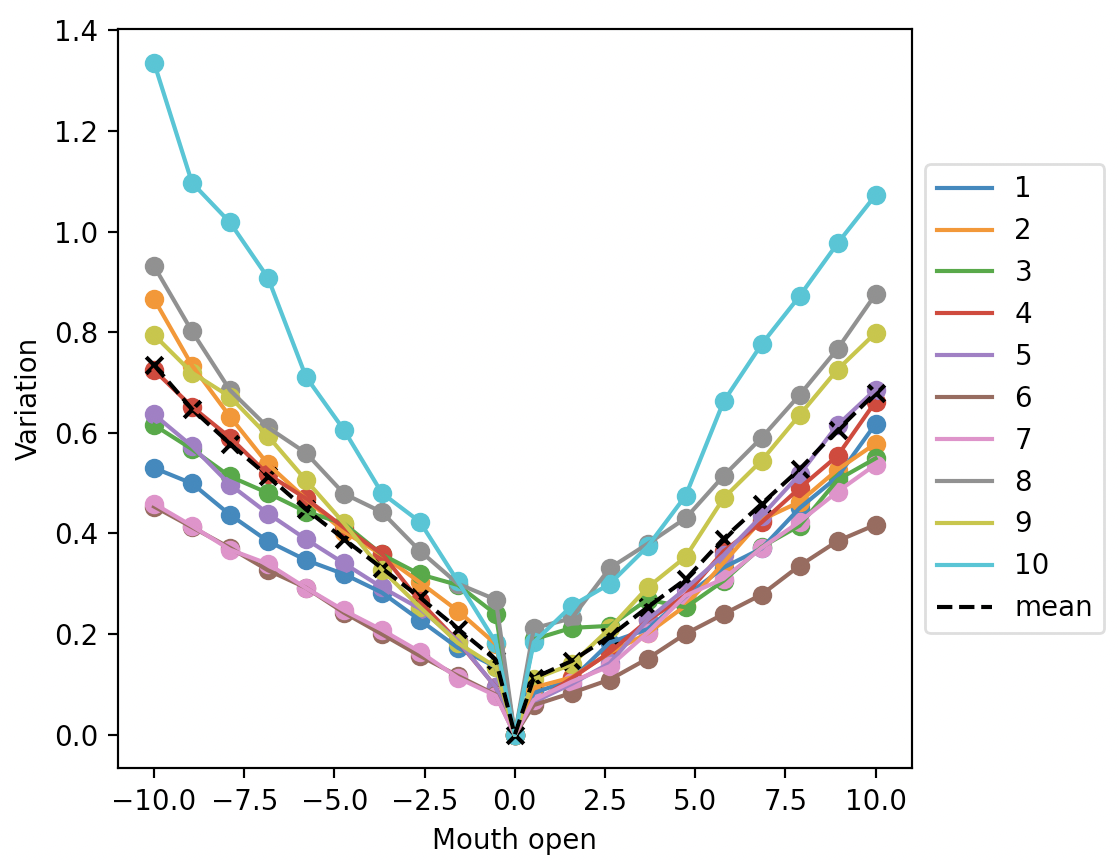}
  \caption{Graph of the face-frame variations through the mouth open direction}
  \label{fig:graph_mouthopen}
\end{figure}

However, figures \ref{fig:graph_mouthopen} and \ref{fig:graph_eyesopen} both show that modifying the openness of the face's mouth or the eyes are the two operations that least increase the face-frame variation.

Another interesting result illustrated in \ref{fig:graph_means} is that all the tested operations have a linear impact on the face-frame variation.

\begin{figure}[H]
  \includegraphics[width=\linewidth]{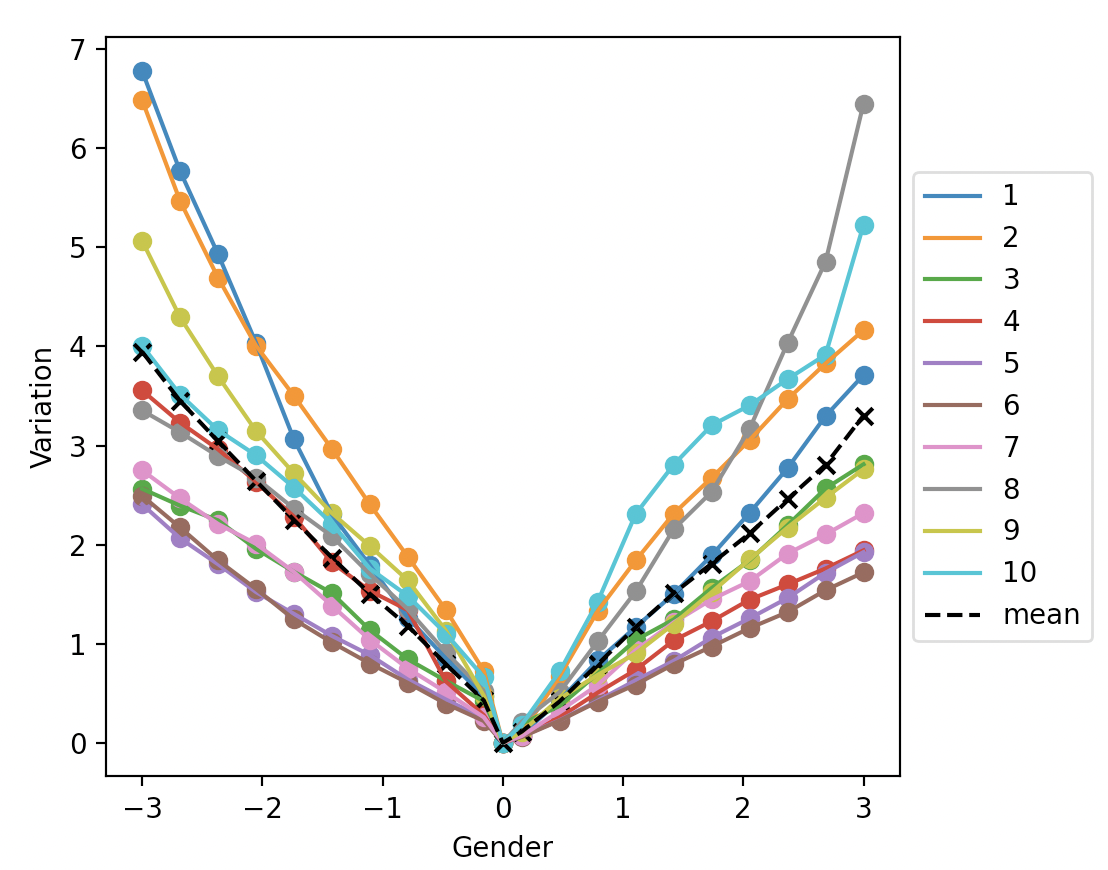}
  \caption{Graph of the face-frame variations through the gender direction}
  \label{fig:graph_gender}
\end{figure}

\biblio % Needed for referencing to working when compiling individual subfiles - Do not remove

%% file: 01_Chapters/07Conclusion.tex
\label{section:conclusion}

In this work, we have explored how different operations in StyleGAN2's latent space affect the face-frame of the resulting facial image. We have also studied how StyleGAN2's image projection affects the face-frame of the original image and we proposed an algorithm to reduce this variation when projecting an image as much as possible.\\
This algorithm can be useful when trying to project a target image generating a new one which is not only similar to the original one in terms of facial characteristics, but it also keeps most the facial frame as much as possible.\\
Besides that, the exploration of the latent space can be useful to understand some underlying aspects of this generative model and the interpretable latent directions. The face-frame variation seems to increase with the magnitude of the movement through any of the studied directions in a linear way. However, the rate of such variation is different for each direction. For example, mouth-open and eyes-open directions presented the least variation while the horizontal and vertical face orientation, age and gender presented the greater variation, which is somewhat expected as the face-frame is different depending on the face's orientation and as the face have different characteristics depending on the age or gender, which may include parts that define its frame.\\
When taking into consideration that the main goal of this paper is to help \emph{Laboratorio of Sueño y Memoria} prove that face-frame plays a key role in the creation of false memories to convict innocent people in identification parades in a crime and that most of the known facts in these crimes are age and gender, these results can be useful for them to project target images faces and modify their relevant features changing as less as possible their frames.

\biblio % Needed for referencing to working when compiling individual subfiles - Do not remove

%% file: 01_Chapters/08FutureWork.tex
\label{section:future_work}
The face-frame correction algorithm and the variation function proposed in this paper provide an insight on how the face-frame may be maintained when a facial image is projected onto the latent space. However, we provide an initial approach which can be further enhanced in order to continue to minimize the face-frame variation while maintaining or even improving the output image fidelity. Here are some ideas we propose:
\subsection{New directions}
Latent directions are the key to modify faces inside the latent space. However, only some of the known latent directions have been studied. New directions can be studied to understand how the latent space is arranged. To find new directions, an unsupervised discovery method can be used, for example the one Voynov and Babenko propose~\citep{unsuperviseddirections}.\\
If one of the found directions is interpretable in a similar way than other known, it is possible to compare them by their face-frame variation to choose between them for a better option.\\
This will allow further face's characteristics modification within the net itself in such a way that it could be possible to help the Lab disprove the belief that police lineups are helpful when searching for a crime perpetrator.
\subsection{Different face-frame variation functions}
The face-frame variation function used for this project described in section \ref{section:function}, is a first version that was created with this project in mind and isn't flawless: sometimes it confuses facial hair, such as beard, as being part of the face and sometimes it labels it as part of the background. The neck is also sometimes considered part of the face, which in reality isn't. There may be a more precise way to measure the face-frame variation between two images that we are not aware of and which can solve this issues.\\
One approach could be taking advantage of facial landmark detection algorithms~\citep{wuandji} to also avoid losing facial characteristics, as our algorithm only considers the face-frame, but it doesn't take into consideration all of the facial characteristics.
\subsection{Improvement of face-frame variation when modifying faces}
The face-frame correction algorithm used is only applied and tested when projecting images to the latent space. However, it may also be used to reduce face-frame variation when performing some operations to a facial image, such as modifying its eye-opening.
\subsection{Different generative networks}
This project was done using StyleGAN2~\citep{stylegan2} as it is the state of art of face images generation at the moment. However, the same experiment can be done using different networks. Text-to-image networks are starting to be popular, it may exist a way that tools such us DALL-E~\citep{dalle} or Stable diffusion~\citep{stablediffusion}, can be used to modify faces with similar results as the ones obtained with StyleGAN2. In fact, DALL-E 2\citep{dalle2} allows for replacing parts of an image using a text description entered in a prompt, in a process called inpainting. This allows for photo realistic modification of images, which yields a credible image in a fast and accurate way.

\begin{figure}[h]
  \includegraphics[width=\linewidth]{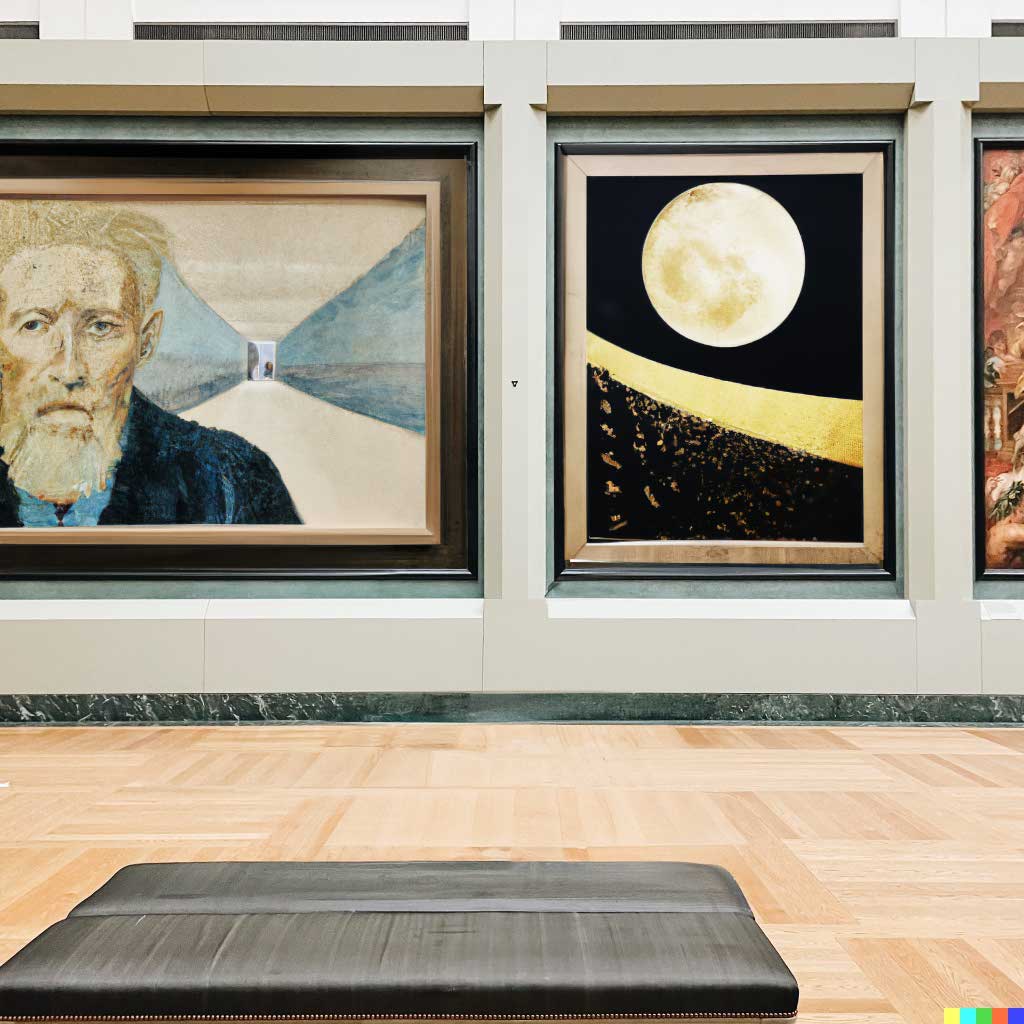}
  \caption{An image of two art pictures in a museum used as an input image to the DALLE 2 AI.}
  \label{fig:dalleinput}
\end{figure}

\begin{figure}[h]
  \includegraphics[width=\linewidth]{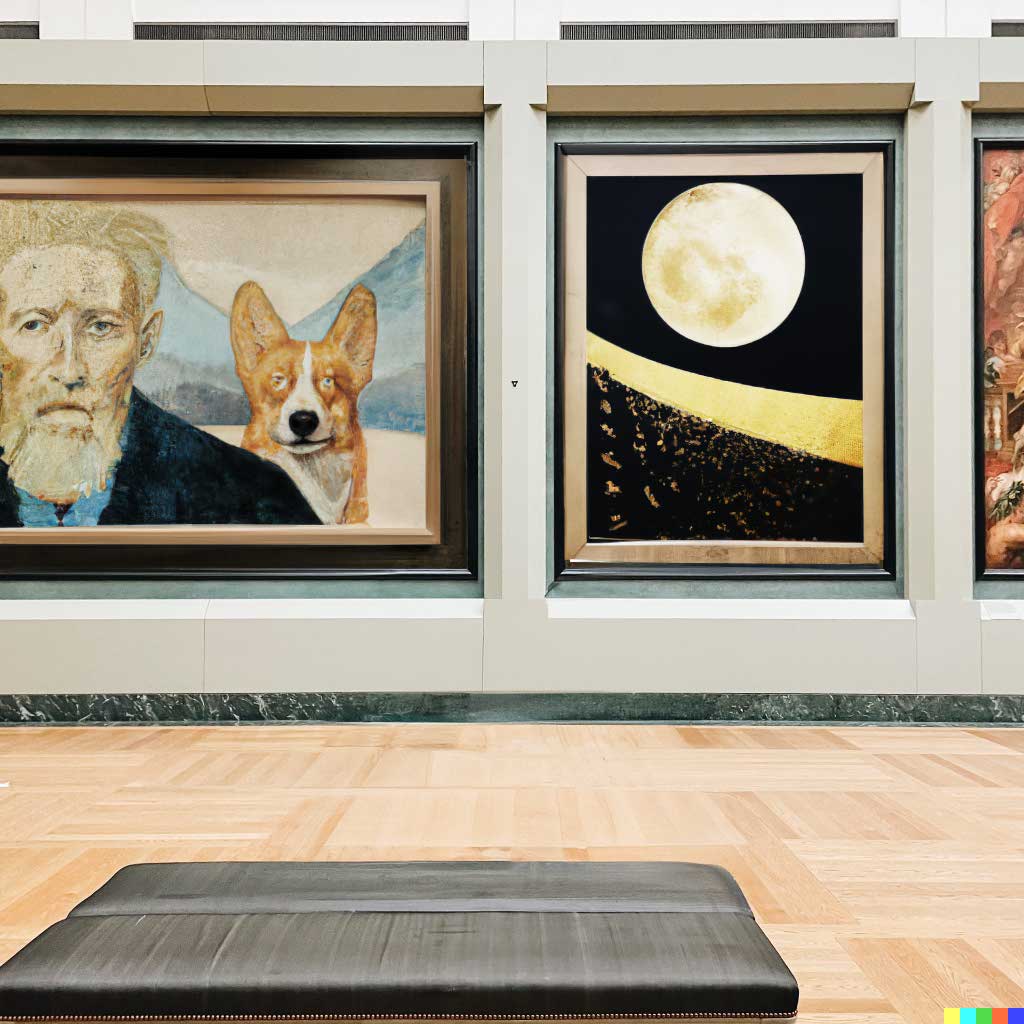}
  \caption{The original museum image including a dog in one of the pictures. This edit was entirely made by DALLE 2 AI.}
  \label{fig:dalle1}
\end{figure}

\begin{figure}[h]
  \includegraphics[width=\linewidth]{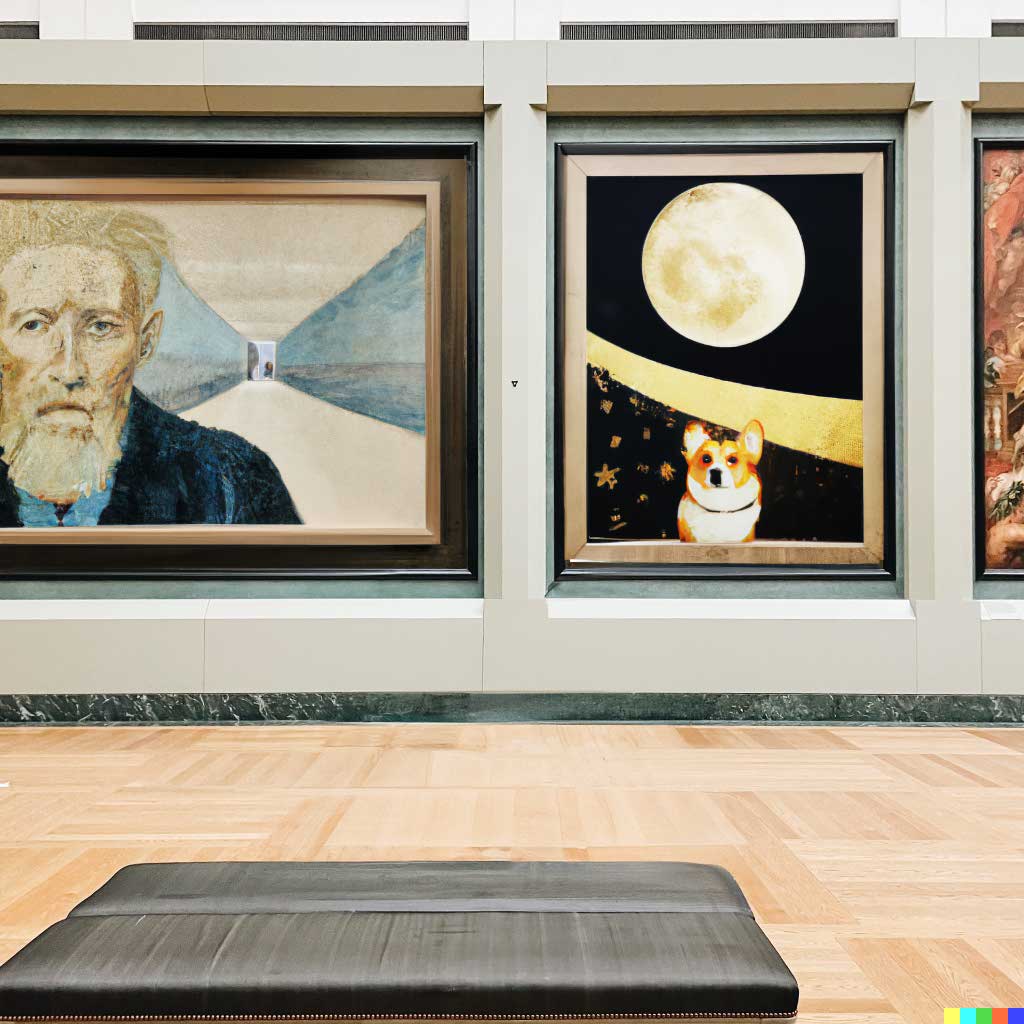}
  \caption{The original museum image including a dog in one of the pictures. This edit was entirely made by DALLE 2 AI.}
  \label{fig:dalle2}
\end{figure}

\begin{figure}[h]
  \includegraphics[width=\linewidth]{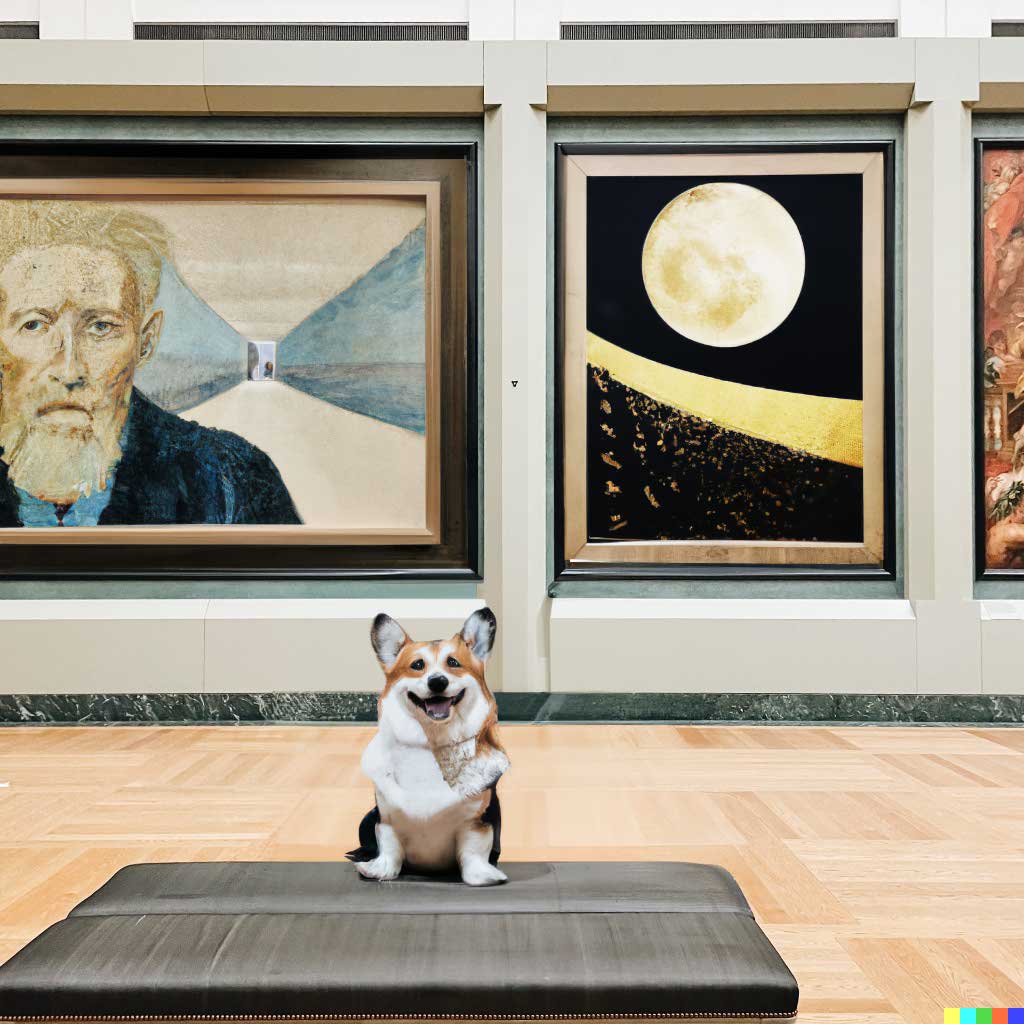}
  \caption{The original museum image including a dog in a sofa. This edit was entirely made by DALLE 2 AI.}
  \label{fig:dalle3}
\end{figure}

\biblio % Needed for referencing to working when compiling individual subfiles - Do not remove

%% file: 01_Chapters/Acknowledgement.tex
\label{section:acknowledgement}

We would like to express our deepest gratitude to our supervisor, Dr. Rodrigo Ramele, who not only presented the topic of this project, but also guided us through it, encouraging us to go deeper than we thought we could. We could not have undertaken this journey without Maite Herrán and Jimena Lozano who paved our way with their project and took the time to explain it to us.\\
Our project would not have been possible without the help of ITBA's IT team who allowed us to make use of their hardware and who set up an environment so that we could use the neural network and perform all the tests we needed. We are also thankful to Dr. Cecilia Forcato and her team in the \emph{Laboratorio de Sueño y Memoria}, who were always helpful and supporting of our project.\\
We thankfully acknowledge the \textbf{NVIDIA Corporation} \emph{Applied Research Accelerator Program} that without their help this project wouldn't have been conducted \#NVIDIAGrant.\\
Lastly, we would like to mention our families, friends and loved ones for their continuous support during this journey.
\biblio % Needed for referencing to working when compiling individual subfiles - Do not remove